%% file: arxiv.tex
\documentclass[10pt,journal,compsoc]{IEEEtran}
\usepackage{times}
\usepackage{epsfig}
\usepackage{graphicx}
\usepackage{amsmath}
\usepackage{amssymb}
\usepackage{listings}
\usepackage[dvipsnames]{xcolor}

\usepackage[utf8]{inputenc} 
\usepackage[T1]{fontenc}    
\usepackage{url}            
\usepackage{booktabs}       
\usepackage{amsfonts}       
\usepackage{nicefrac}       
\usepackage{microtype}      

\usepackage{graphicx}
\usepackage{multirow}
\usepackage{caption}
\usepackage{amsmath}
\usepackage{amssymb}
\usepackage{xspace}
\usepackage{algorithm,algpseudocode}
\usepackage{mmstyle}
\usepackage{color}
\usepackage{subcaption}
\usepackage{tikz}
\usepackage{pgfplots}

\usepackage{stfloats}

\renewcommand{\paragraph}[1]{\noindent\textbf{#1}~~}
\usepackage[pagebackref=false,breaklinks=true,colorlinks,bookmarks=false]{hyperref}

\definecolor{codeblue}{rgb}{0.25,0.5,0.5}

\algnewcommand\algorithmicinput{\textbf{Input:}}
\algnewcommand\INPUT{\item[\algorithmicinput]}
\algnewcommand\algorithmicinputt{\textbf{Track:}}
\algnewcommand\TRACK{\item[\algorithmicinputt]}
\newcommand{\LineComment}[1]{\hfill \textcolor{codeblue}{\# #1}}

\ifCLASSOPTIONcompsoc
  \usepackage[nocompress]{cite}
\else
  \usepackage{cite}
\fi
\ifCLASSINFOpdf
\else
\fi
\begin{document}
%
\title{QDTrack: Quasi-Dense Similarity Learning for Appearance-Only Multiple Object Tracking}
%
%
%
%

\author{
Tobias Fischer*,
Thomas E. Huang*,
Jiangmiao Pang*,
Linlu Qiu,
Haofeng Chen,
Trevor Darrell,
Fisher Yu
\IEEEcompsocitemizethanks{
\IEEEcompsocthanksitem T. Fischer, T. E. Huang, F. Yu are with the Department of Information Technology and Electrical Engineering, ETH Zurich.\protect\\
E-mail: tobias.fischer@vision.ee.ethz.ch
\IEEEcompsocthanksitem J. Pang is with the Shanghai AI Laboratory.
\IEEEcompsocthanksitem L. Qiu is with the Department of EECS, Massachusetts Institute of Technology.
\IEEEcompsocthanksitem H. Chen is with the Computer Science Department, Stanford University.
\IEEEcompsocthanksitem T. Darrell is with the Department of EECS, UC Berkeley.
\\ * Equal contribution.
}
}

\IEEEtitleabstractindextext{%
\input{sections/abstract}

}

\maketitle

\IEEEdisplaynontitleabstractindextext

%
\IEEEpeerreviewmaketitle

\input{sections/introduction}
\input{sections/related_work}
\input{sections/method}
\input{sections/experiments}
\input{sections/conclusions}



\ifCLASSOPTIONcaptionsoff
  \newpage
\fi



%



{\small
    \bibliographystyle{IEEEtran}
    \bibliography{egbib}
}


\input{sections/appendix}
\end{document}

%% file: sections/abstract.tex

\begin{abstract}

Similarity learning has been recognized as a crucial step for object tracking.
However, existing multiple object tracking methods only use sparse ground truth matching as the training objective, while ignoring the majority of the informative regions in images.
In this paper, we present Quasi-Dense Similarity Learning, which densely samples hundreds of object regions on a pair of images for contrastive learning.
We combine this similarity learning with multiple existing object detectors to build Quasi-Dense Tracking (QDTrack), which does not require displacement regression or motion priors.
We find that the resulting distinctive feature space admits a simple nearest neighbor search at inference time for object association. 
In addition, we show that our similarity learning scheme is not limited to video data, but can learn effective instance similarity even from static input, enabling a competitive tracking performance without training on videos or using tracking supervision.
We conduct extensive experiments on a wide variety of popular MOT benchmarks.
We find that, despite its simplicity, QDTrack rivals the performance of state-of-the-art tracking methods on all benchmarks and sets a new state-of-the-art on the large-scale BDD100K MOT benchmark, while introducing negligible computational overhead to the detector.


\end{abstract}

%% file: sections/introduction.tex
\IEEEraisesectionheading{\section{Introduction}}
\IEEEPARstart{M}{ultiple} Object Tracking (MOT) is a fundamental and challenging problem in computer vision, widely used in safety monitoring, autonomous driving, video analytics, and other applications.
Contemporary MOT methods~\cite{tracktor, sort, deepsort, centertrack, jde} mainly follow the tracking-by-detection paradigm~\cite{ramanan2003finding}. That is, they detect objects on each frame and then associate them according to the estimated similarity between each instance.
Recent works~\cite{tracktor, sort, ioutracker, centertrack} show that if the detected objects are accurate,
the spatial proximity between objects in consecutive frames, measured by Intersection over Union (IoU) or center distance, is a strong prior to associate the objects.
However, this location prior is often violated in more complex scenarios with non-linear object motion, varying video frame rate, or complex camera motion, since the movement of objects on the image plane depends highly on these factors.
To remedy this problem, some methods introduce motion estimation~\cite{andriyenko2011multi, choi2010multiple} or displacement regression~\cite{d&t,centertrack,chainedtracker} to ensure accurate distance estimation.
Object appearance similarity usually takes a secondary role~\cite{retinatrack, deepsort} to strengthen object association or re-identify vanished objects, because extracted appearance features cannot effectively distinguish different objects.
Thus, the search region is constrained to local neighborhoods to avoid distractions.

On the contrary, humans can easily associate identical objects only through appearance.
We conjecture this is because the image and object information is not fully utilized for learning object similarity. Previous methods regard instance similarity learning as a post-hoc stage after object detection or only use sparse ground truth bounding boxes as training samples~\cite{deepsort}.
These processes ignore the majority of the informative regions on the images. 
We hypothesize that, because objects in an image are rarely identical to each other, a nearest neighbor search in a learned feature space should associate and distinguish instances without bells and whistles.
In addition, we observe that besides the ground truth and detected bounding boxes, which sparsely distribute on the images, many possible object regions can provide valuable training supervision.

In this paper, we propose quasi-dense similarity learning, which densely matches hundreds of informative regions on a pair of images for contrastive learning.
The quasi-dense samples cover a wide range of informative regions on the images, providing both more positive examples and hard negatives.
Because one sample has more than one positive counterpart on the reference image, we extend the InfoNCE loss~\cite{oord2018representation} commonly used in contrastive learning~\cite{hadsell2006dimensionality, sohn2016improved, wu2018unsupervised} to multiple positives which makes quasi-dense learning feasible.
Each sample is thus trained to discriminate an instance from all possible object regions on the reference image simultaneously.
This provides stronger supervision than using only a handful ground truth labels and enhances the instance similarity learning.
To extract feature embeddings for each region, we use a lightweight embedding extractor that works with most existing object detectors.

Besides similarity, the inference pipeline, which measures the instance similarity and maintains a track history, also plays an important role in the tracking performance, since it needs to consider false positives, missed detections, newly appeared objects, and terminated tracks.
To better deal with these cases, we introduce the \emph{bi-directional softmax} similarity metric that enforces bi-directional consistency.
In particular, objects that do not have matching targets in the other frame will lack a bi-directional matching and thus have low similarity scores to all other objects.
Furthermore, we include unmatched objects in the previous frame, which we call backdrops, for matching to better filter false positives that could otherwise act as distractors in following frames.
We compose object detectors, quasi-dense similarity learning, and our inference pipeline to build \emph{Quasi-Dense Tracking} (QDTrack) models.
Since the publication of our initial work~\cite{qdtrack}, QDTrack has been widely adopted for other tracking problems, such as segmentation tracking~\cite{ke2021prototypical}, long-tailed multi-object tracking~\cite{trackeverything}, and 3D object tracking~\cite{hu2022monocular}.

In addition to the findings of our initial work~\cite{qdtrack}, we show that quasi-dense instance similarity learning is not limited to video data, but can learn effective instance representations from static images alone. In particular, we show that we can effectively perform tracking even when learning instance similarity without any annotations for association and/or video input.
Moreover, in this journal extension we conduct extensive experiments on a wide variety of tracking benchmarks, namely MOT~\cite{mot16}, DanceTrack~\cite{peize2021dance}, BDD100K~\cite{bdd100k}, Waymo~\cite{waymo}, and TAO~\cite{tao}. In addition, we show the flexibility of our method by combining it with different base models and object detectors.
Despite its simplicity, QDTrack rivals the performance of state-of-the-art methods without bells and whistles, and sets a new state-of-the-art on the large-scale BDD100K tracking benchmark. 
QDTrack allows for joint, end-to-end training of detection and instance similarity, thereby simplifying the training and inference pipelines of MOT frameworks. In addition, our embedding extractor only adds negligible overhead to the inference time of the detector.
We hope the simplicity and strengths of QDTrack motivates further research on similarity learning for multiple object tracking.

%% file: sections/related_work.tex

\begin{figure*}
  \centering
  \includegraphics[width=\linewidth]{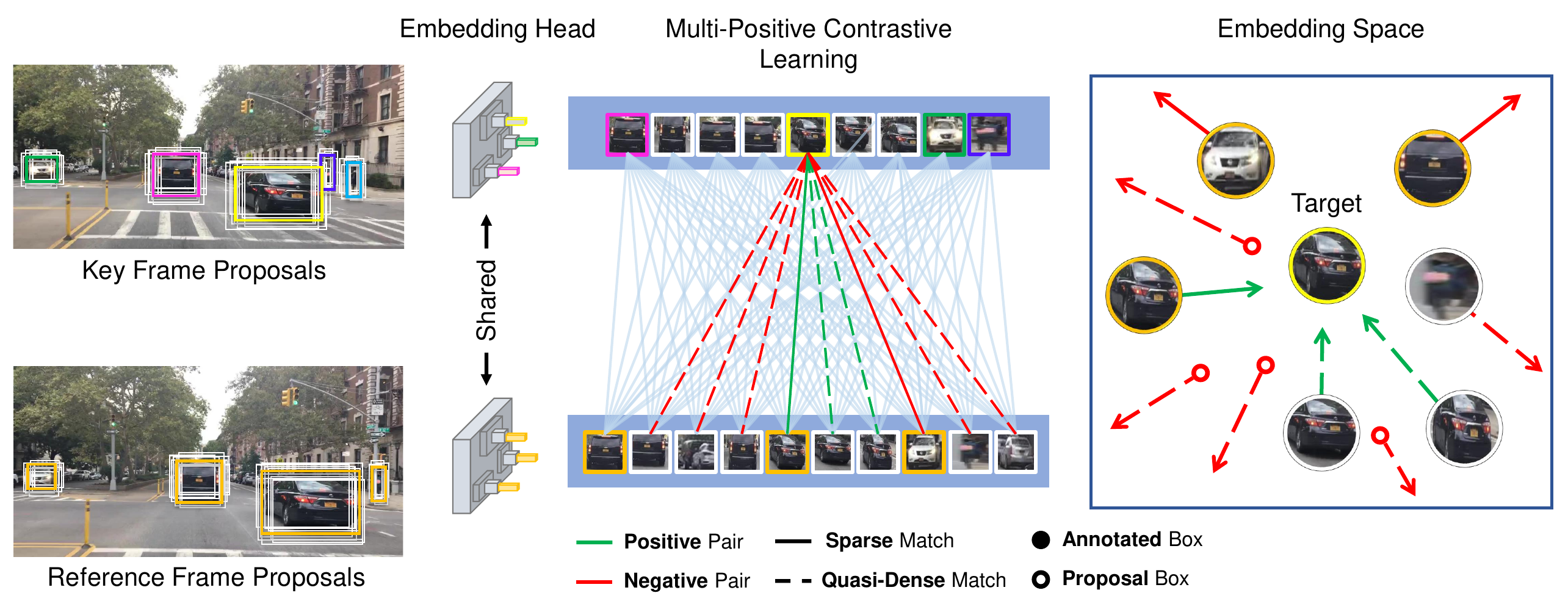}
  \caption{\textbf{Training pipeline}. After we extract feature embeddings for all quasi-dense samples on a pair of key and reference images, we apply dense matching between them and optimize the learned representation with multiple positive contrastive learning. The resulting embedding space effectively discriminates different instances.}
  \label{fig:train}
\end{figure*}

\section{Related work}
In MOT, the current leading paradigm is tracking-by-detection~\cite{ramanan2003finding}.
Tracking-by-detection methods detect objects in each individual frame, and subsequently associate the detections over time. They differ in their data association mechanisms and cues that are used in the association process.
A variety of approaches have been developed to solve the data association problem, \eg, network flow formulations~\cite{zhang2008}, quadratic pseudo boolean optimization~\cite{ess2008mobile}, conditional random fields~\cite{nomt}, or multi-hypothesis tracking~\cite{kim2015multiple}.
Many works have focused on finding the best cues to exploit for data association, such as 2D motion~\cite{sort, ioutracker, xiao2018simple, poi, goturn, d&t}, 3D motion~\cite{osep2017combined, beyondpixels, mitzel2012taking, motsfusion, hu2022monocular},
or visual appearance similarity~\cite{chanho2015, laura2016, crf, jeany2017, amir2017, anton2017, deepsort, tracktor}.
In this work, we focus on learning visual appearance similarity and follow the tracking-by-detection paradigm.

\paragraph{Location and motion in MOT}
Spatial proximity has been proven effective to associate objects in consecutive frames~\cite{sort, ioutracker}.
Some methods use 2D motion, such as predictions of a Kalman Filter~\cite{sort, poi, zhang2021bytetrack, cao2022observation}, optical flow~\cite{xiao2018simple}, and displacement regression~\cite{goturn, d&t}, to estimate similarity for object association.
However, these methods are brittle when it comes to varying video frame rate and complex camera motion, since the 2D motion of the objects depends highly on these factors.
%
Thus, other methods instead rely on 3D motion cues to associate objects over time, since in 3D camera and object motion can be decomposed.
A common paradigm~\cite{osep2017combined, beyondpixels} is to track objects with 3D bounding
boxes and motion estimates derived from \eg scene flow.
In contrast, \cite{held2013precision, mitzel2012taking, motsfusion} explored to track and reconstruct objects in 3D by estimating the object's rigid-body transformation between two frames.
Although these methods show promising results, many \cite{deepsort, zhang2021bytetrack} still rely on an extra appearance similarity model as a complementary component to re-identify vanished objects, complicating the entire framework.
Our approach is orthogonal to this line of work, as we rely solely on appearance-based instance similarity and a simple nearest-neighbor search to associate objects.

\paragraph{Appearance similarity in MOT}
In order to strengthen tracking and re-identify vanished objects, some methods exploit appearance similarity extracted from an independent model~\cite{chanho2015, laura2016, crf, jeany2017, amir2017, anton2017, deepsort, tracktor} or add an extra embedding head to the detector for end-to-end training~\cite{trackrcnn, retinatrack, jde, zhang2020fair}.
However, they still learn appearance similarity following the practice in image similarity learning, and measure the instance similarity by cosine distance.
An appearance similarity model is trained either as a $n$-class classification problem~\cite{deepsort}, where $n$ is equal to the number of identities in the whole training set, or using triplet loss~\cite{tripletloss}.
The classification problem is hard to extend to large-scale datasets, while the triplet loss only compares each training sample with two other samples, leading to sub-optimal instance similarity learning.
As a consequence, these methods still rely heavily on motion models and displacement predictions to track objects, and appearance similarity only takes a secondary role.
In contrast, QDTrack learns instance similarity from densely-connected contrastive pairs and associates objects with a simple nearest neighbor search in feature space, which allows for a simpler tracking framework without compromising accuracy. 

\paragraph{Joint detection and tracking}
Instead of treating object detection and association as separate modules, Detect \& Track~\cite{d&t} is the first work that jointly optimizes object detection and tracking modules.
It predicts the displacements of each object in consecutive frames and associates them with the Viterbi algorithm.
Tracktor~\cite{tracktor} directly adopts a detector as a tracker. 
CenterTrack~\cite{centertrack} and Chained-Tracker~\cite{chainedtracker} predict the object displacements with pair-wise inputs to associate the objects.
Other methods focus on learning visual appearance and detection jointly~\cite{trackrcnn, zhang2020fair, jde}, adding an extra embedding head to the detection network.
However, these methods do not fully exploit image information for similarity learning.
Recent work~\cite{meinhardt2021trackformer, zeng2021motr, transtrack} focuses on leveraging Transformer networks to integrate tracking and detection into a single, query-based architecture.
These methods track by propagating queries across timesteps, processing them with a Transformer that outputs the tracking result.
In this work, we focus on learning appearance similarity from quasi-dense samples jointly with detection.

\paragraph{Self-supervised representation learning}
The field of self-supervised representation learning has seen significant progress in recent years, fueled by a number of methods relying on contrastive learning~\cite{bachman2019learning, henaff2019data, oord2018representation, tian2019contrastive, wu2018unsupervised, he2020momentum, chen2020simple, circleloss} that have shown promising performance. 
The main paradigm of these methods is to learn a representation that is similar for two versions of the same image, where one is distorted with random image augmentations, while enforcing that this representation is dissimilar to other pairs in the current training batch.
While this has proven to be very effective, it has not yet drawn much attention when learning the instance similarity in MOT.
In this paper, we supervise densely matched quasi-dense samples with multiple positive contrastive learning inspired by~\cite{circleloss}.
In contrast to image-level contrastive learning, our method allows for multiple positive training, while the methods mentioned above can only handle the case when there is only a single positive target.
The promising results of our method show the importance of representation learning for the MOT problem.

\paragraph{Learning to track from static images}
Learning to track objects from static images where no association annotations are available has recently been proposed by multiple methods~\cite{zhang2020fair, centertrack}. CenterTrack~\cite{centertrack} proposes to use data augmentation to simulate video input from given a single static image to obtain 2D offsets to learn object motion.
FairMOT~\cite{zhang2020fair} treats every object in a given detection dataset as a unique class and learns to distinguish between those to learn tracking from static images.
In contrast to learning simulated motion or treating every object over a whole dataset as unique, we show that our similarity learning scheme can effectively learn to track objects from static images with comparable accuracy to video input without further modification.
We draw inspiration from the success of recent self-supervised representation learning methods and apply our similarity learning scheme between two augmented instances of the same input image.

%% file: sections/method.tex

\section{Method}

We propose \emph{quasi-dense similarity learning} to learn a feature embedding space that can associate identical objects and distinguish different objects for online multiple object tracking.
We define \emph{dense matching} to be matching between bounding box candidates at all pixel locations and \emph{sparse matching} to be matching between ground truth box labels as matching candidates.
In contrast, \emph{quasi-dense matching} considers potential object candidates specifically at potential object regions.
The main ingredients of \textit{Quasi-Dense Tracking} (QDTrack) are object detection, instance similarity learning, and object association.

\subsection{Object detection}
Our method can be easily coupled with both two-stage and one-stage detectors with end-to-end training.
Object detectors contain two components, a feature extractor and a bounding box prediction head.
The feature extractor is typically composed of a base model to extract image features and a Feature Pyramid Network (FPN)~\cite{fpn} to obtain multi-scale features.
The bounding box prediction head produces dense bounding box candidates, from which we sample quasi-dense samples by filtering with Non-Maximal Suppression (NMS).
The resulting samples indicate likely object regions that include multiple overlapped bounding boxes for each object.

\subsection{Quasi-dense similarity learning} \label{sec:quasi-dense}
We use regions that likely contain objects to learn the instance similarity with quasi-dense matching.
The full training pipeline is
shown in Figure~\ref{fig:train}.
Given a key image $I_1$ for training, we randomly select a reference image $I_2$ from its temporal neighborhood.
The neighbor distance is constrained by an interval $k$.
We use the object regions from both images and RoI Align~\cite{maskrcnn} to obtain their corresponding feature maps from the image features.
We add an extra lightweight embedding head in parallel with the original bounding box head to extract features for each region.
A region is defined as a positive sample to a ground truth object if it has an IoU higher than $\alpha_1$ or negative if lower than $\alpha_2$.
A matching of regions on two frames is positive if the two regions are associated with the same ground truth object and negative otherwise.

\begin{figure*}[t]
    \centering
    \includegraphics[width=\linewidth]{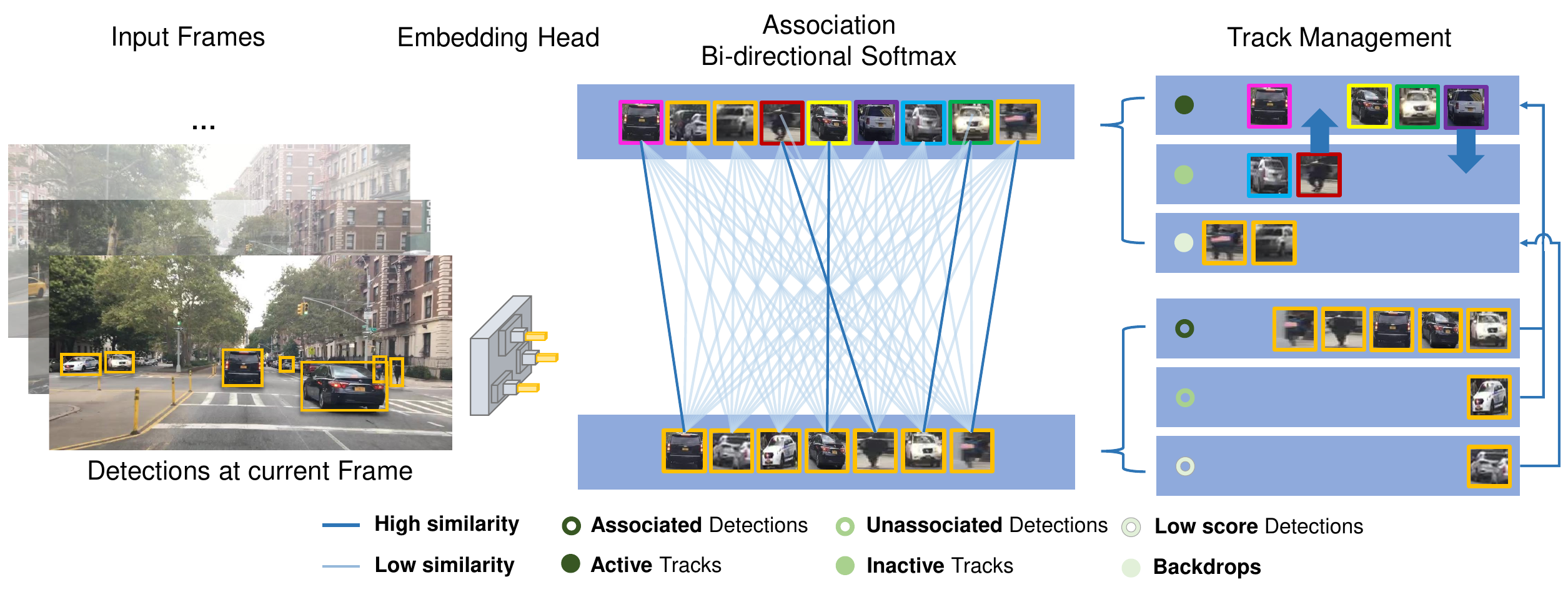}
    \caption{\textbf{Inference pipeline}. First, we extract object detections and their corresponding feature embeddings from the current frame. Next, we use bi-softmax to measure the instance similarity between all detections and matching candidates. Finally, we associate objects with a simple nearest neighbor search in the feature space and update our track history.}
    \label{fig:test}
\end{figure*}

Assume there are $V$ samples on the key frame as training samples and $K$ samples on the reference frame as contrastive targets.
For each training sample, we can use the non-parametric softmax~\cite{wu2018unsupervised, oord2018representation} with cross-entropy to optimize the feature embeddings,
\begin{align}
    \mathcal{L}_\text{embed} & = -\text{log}
    \frac{\text{exp}(\textbf{v} \cdot \textbf{k}^{+})}
    {\text{exp}(\textbf{v} \cdot \textbf{k}^{+}) + \sum_{\textbf{k}^{-}}\text{exp}(\textbf{v} \cdot \textbf{k}^{-})},
    \label{eqa:npair}
\end{align}
where $\textbf{v}$, $\textbf{k}^{+}$, $\textbf{k}^{-}$ are feature embeddings of the training sample, its positive target, and negative targets in $K$.
The overall embedding loss is averaged across all training samples, but we only illustrate one training sample for brevity.

We apply dense matching between object regions on the pairs of images. Specifically, each sample in $I_1$ is matched to all samples in $I_2$, in contrast to only using sparse sample crops (mostly ground truth boxes) to learn instance similarity in previous works~\cite{bertinetto2016fully, tripletloss}.
Each training sample in the key frame has more than one positive targets in the reference frame, so Eq.~\eqref{eqa:npair} can be extended as
\begin{align}
    \mathcal{L}_\text{embed} & = -\sum_{\textbf{k}^{+}}\text{log}
    \frac{\text{exp}(\textbf{v} \cdot \textbf{k}^{+})}
    {\text{exp}(\textbf{v} \cdot \textbf{k}^{+}) + \sum_{\textbf{k}^{-}}\text{exp}(\textbf{v} \cdot \textbf{k}^{-})}.
    \label{eqa:npair2}
\end{align}

\begin{algorithm}[t]
    \caption{Inference pipeline of QDTrack for associating objects across a video sequence.}
    \label{alg:inference}
    \begin{algorithmic}[1]
        \INPUT frame index $t$, detections $\textbf{b}_i$, scores $s_i$, detection embeddings $\textbf{n}_i$ for $i=1 \ldots N$, and track embeddings $\textbf{m}_j$ for $j=1 \ldots M$
        \State \texttt{DuplicateRemoval}($\textbf{b}_i$)
        \For{$i=1\ldots N, j=1\ldots M$} \LineComment{compute matching scores}
        \State \textbf{f}$(i, j) = $ \texttt{bisoftmax}($\textbf{n}_i, \textbf{m}_j$)
        \EndFor
        \For{$i=1\ldots N$} \LineComment{track management}
        \State $c$ = \texttt{max}$\left(\textbf{f}(i)\right)$ \LineComment{match confidence}
        \State $j_{\texttt{match}}$ = \texttt{argmax}$\left(\textbf{f}(i)\right)$ \LineComment{matched track ID}
        \If{$c$ > $\beta_{\texttt{match}}$ \textbf{and} $s_i >$ $\beta_{\texttt{obj}}$}
        \Statex \hskip1.5em \textbf{and} \texttt{isNotBackdrop}$\left(j_{\texttt{match}}\right)$ \LineComment{object match found}
        \State \texttt{updateTrack}$\left(j_{\texttt{match}}, \textbf{b}_i, \textbf{n}_i, t\right)$  \LineComment{update track}
        \ElsIf{$s_i >$ $\beta_{\texttt{new}}$}
        \State \texttt{createTrack}$\left(\textbf{b}_i, \textbf{n}_i, t\right)$ \LineComment{create new track}
        \Else{}
        \State \texttt{addBackdrop}$\left(\textbf{b}_i, \textbf{n}_i, t \right)$ \LineComment{add new backdrop}
        \EndIf
        \EndFor
    \end{algorithmic}
\end{algorithm}

However, this equation does not treat positive and negative targets fairly. Namely, each negative is considered multiple times while each positive is considered only once.
Alternatively, we can first reformulate Eq.~\eqref{eqa:npair} as
\begin{align}
    \mathcal{L}_\text{embed}
     & = \text{log}\left[1 + \sum_{\textbf{k}^{-}} \text{exp}(\textbf{v} \cdot \textbf{k}^{-} - \textbf{v} \cdot \textbf{k}^{+})\right].
\end{align}
Then in the multi-positive scenario, it can be extended by accumulating the positive term as
\begin{align}
    \mathcal{L}_\text{embed} =
    \text{log}\left[1 + \sum_{\textbf{k}^{+}} \sum_{\textbf{k}^{-}} \text{exp}(\textbf{v} \cdot \textbf{k}^{-} - \textbf{v} \cdot \textbf{k}^{+})\right].
    \label{eqa:multipos}
\end{align}
We further adopt L2 loss as an auxiliary loss
\begin{equation}
    \mathcal{L}_\text{aux} = \left(\frac{\textbf{v} \cdot \textbf{k}}{||\textbf{v}|| ||\textbf{k}||} - c\right)^2,
\end{equation}
where $c$ is 1 if the match of two samples is positive and 0 otherwise.
Note the auxiliary loss aims to constrain the magnitude of the logits and cosine similarity instead of improving the performance. We sample all positive pairs and three times more negative pairs to calculate the auxiliary loss and use hard negative mining.

The entire network is jointly optimized under
\begin{equation}
\label{eqa:loss}
    \mathcal{L} = \mathcal{L}_\text{det} + \gamma_1 \mathcal{L}_\text{embed} + \gamma_2 \mathcal{L}_\text{aux},
\end{equation}
where $\gamma_1$ and $\gamma_2$ are set to 0.25 and 1.0 by default in this paper.

\subsection{Object association and track management}
Tracking objects across frames purely based on object feature embeddings introduces many challenges.
False positives, ID switches, newly appeared objects, and terminated tracks all increase the matching difficulty.
We here introduce our inference pipeline that utilizes instance similarity for object association and a track management scheme to address these problems.
The entire pipeline is shown in Figure~\ref{fig:test} and described in Algorithm~\ref{alg:inference}.


\paragraph{Duplicate removal}
Most object detectors only use intra-class NMS to remove duplicate detections within each class, which results in some detections that are in the same location but with different categories.
For object tracking, this is undesirable as it will create duplicate object embeddings.
We instead use inter-class NMS to avoid this issue.

\paragraph{Bi-directional softmax}
Our main inference strategy is bi-directional matching in the feature embedding space.
Assume there are $N$ detected objects in frame $t$ with feature embeddings $\textbf{n}$ and $M$ matching candidates with feature embeddings $\textbf{m}$ from the past $x$ frames. The instance similarity $\textbf{f}$ between objects and their matching candidates is obtained by a bi-directional softmax (bi-softmax):
\begin{equation}
    \textbf{f}(i, j) = \frac{1}{2} \left[\frac{ \text{exp}(\textbf{n}_i \cdot \textbf{m}_j)}{\sum_{k=0}^{M-1} \text{exp}(\textbf{n}_i \cdot \textbf{m}_k )} +
    \frac{\text{exp}(\textbf{n}_i \cdot \textbf{m}_j)}{\sum_{k=0}^{N-1} \text{exp}(\textbf{n}_k \cdot \textbf{m}_j )}\right].
\end{equation}
A high score under bi-softmax indicates that the two matched objects are each other's nearest neighbor in the feature space, thus satisfying bi-directional consistency.
$\textbf{f}$ can be used to directly associate objects with a simple nearest neighbor search.

\paragraph{Track management}
We use a track management scheme to keep track of inactive and currently active tracks and to handle the matching of objects.
Active tracks are tracks that have a matching detection in the previous frame, otherwise they become inactive.
Tracks that are inactivate for $K$ frames will be removed and not be considered for matching.
Detections are only considered for matching to existing tracks if the detection confidence is above a threshold $\beta_{\texttt{obj}}$.
A match is determined if the matching score is higher than a threshold $\beta_{\texttt{match}}$.

Objects without a target in the feature space should not be matched to any candidates.
Newly appeared objects, vanished tracks, and some false positives fall into this category.
Bi-softmax can handle such objects, as it is difficult to achieve high matching scores in both directions due to the uncertainty in matching.
Thus, these objects will likely obtain a low bi-softmax score and will not be matched to any existing tracks.
For such objects that have a detection confidence higher than a threshold $\beta_{\texttt{new}}$, we initialize a new track instead.

Most detections with low confidence that do not match any existing tracks are false positives that introduce uncertainty to the matching process.
Previous methods often directly drop them and do not consider them again.
We argue that these false positives appear frequently in the following frames, which hurts tracking performance.
To remedy this, we keep the unmatched objects as \emph{backdrops} for $L$ frames and use them as matching candidates.
Detections that are matched to backdrops will thus not be matched to existing tracks.
Our experiments show that backdrops can reduce the number of false positives.




%% file: sections/experiments.tex

\section{Experiments}
We conduct experiments on a variety of MOT benchmarks including MOT17~\cite{mot16} and MOT20~\cite{dendorfer2020mot20}, DanceTrack~\cite{peize2021dance}, BDD100K~\cite{bdd100k}, Waymo~\cite{waymo}, and TAO~\cite{tao}, and compare our method extensively to the state-of-the-art.
In addition, we show that we can effectively perform tracking even without tracking supervision or video data.
We demonstrate the flexibility of our method by combining it with different detection methods and feature-extraction base models and conduct extensive ablation studies on all aspects of our method. 
Finally, we also present a straightforward extension of our method to segmentation tracking and give insights on the limitations of our method.
More detailed oracle and failure case analyses are presented in the appendix.

\subsection{Datasets}

\paragraph{MOT Challenge}
We perform experiments on two of the MOT Challenge benchmarks, namely MOT17~\cite{mot16} and MOT20~\cite{dendorfer2020mot20}. The MOT Challenge videos contain high-density public spaces such as street scenes and malls with many pedestrians, creating challenging tracking conditions with heavy occlusions. Only pedestrians are evaluated in this benchmark. Since these datasets do not provide official validation sets, we split each training video into two halves: the first half for training and the second half for validation following~\cite{centertrack, jde, zhang2020fair}.

The MOT17 dataset contains 7 videos (5,316 images) for training and 7 videos (5,919 images) for testing. The video frame rate is 25 - 30 FPS.
The MOT20 dataset includes heavily crowded scenes and contains 4 videos (8,931 images) for training and 4 videos (4,479 images) for testing. The video frame rate is 25 FPS.

\paragraph{DanceTrack}
The DanceTrack \cite{peize2021dance} benchmark is a large-scale dataset for multi-human tracking consisting mostly of group dancing videos. The dataset is unique in that by relying mostly on group dancing videos, the objects to track often have similar appearance, diverse motion, and extreme articulation. It features 40 videos for training, 25 videos for validation and 35 videos for testing, with a total of 105,855 frames captured at 20 FPS.

\paragraph{BDD100K}
The large-scale, diverse driving dataset BDD100K~\cite{bdd100k} contains 100,000 video sequences of dashcam driving footage. It contains several subsets with different types of annotations. We use the detection and tracking sets for training and the tracking set for evaluation. The tracking set annotates 8 categories for evaluation. It contains 1,400 videos (278k images) for training, 200 videos (40k images) for validation, and 400 videos (80k images) for testing. The detection set has 70,000 images for training. The images in the tracking set are annotated at 5 FPS.
\input{tables/comparison_mot}

\paragraph{Waymo}
Waymo open dataset~\cite{waymo} contains images from 5 cameras associated with 5 different directions: front, front left, front right, side left, and side right.
There are 3,990 videos (790k images) for training, 1,010 videos (200k images) for validation, and 750 videos (148k images) for testing.
It annotates 3 classes for evaluation.
The videos are annotated at 10 FPS.

\input{tables/comparison_dancetrack}

\input{tables/comparison_bdd100k}

\input{tables/comparison_waymo}

\begin{table}[t]
    \caption{\textbf{Comparison to state-of-the-art on TAO.} We evaluate and compare our method on the TAO challenge benchmark. We use Faster R-CNN~\cite{frcnn} as our detection method. $\dagger$ indicates offline methods, $\ddagger$ indicates methods using additional data.}
    \centering
    \resizebox{\linewidth}{!}{
        \begin{tabular}{clcccccc}
            \toprule
            Split              &  Method     & AP50 & AP75 & AP  & AP50(S) & AP50(M) & AP50(L) \\
            \midrule
            \multirow{4}{*}{val} & SORT\_TAO~\cite{tao}  & 13.2 & -    & -   & -       & -       & -       \\
            & QDTrack (Ours)                  & 16.1 & 5.0  & 7.0 & 2.4     & 4.6     & 9.6     \\ \cmidrule{2-8}
            & \textcolor{gray}{GTR}~\cite{zhou2022global}~$\dagger$                  & \textcolor{gray}{22.5} & -  & - & -     & -     & - \\
            & \textcolor{gray}{AOA}~\cite{du2021aoa}~$\ddagger$                  & \textcolor{gray}{25.8} & -  & - & -     & -     & - \\
            \midrule
            \multirow{4}{*}{test} & SORT\_TAO~\cite{tao} & 10.2 & 4.4  & 4.9 & 7.7     & 8.2     & 15.2    \\
             & QDTrack (Ours)      & 12.4 & 4.5  & 5.2 & 3.7     & 8.3     & 18.8    \\ \cmidrule{2-8}
             & \textcolor{gray}{GTR}~\cite{zhou2022global}~$\dagger$              & \textcolor{gray}{20.1} & -  & - & -     & -     & - \\
             & \textcolor{gray}{AOA}~\cite{du2021aoa}~$\ddagger$              & \textcolor{gray}{27.5} & -  & - & -     & -     & - \\
            \bottomrule
        \end{tabular}
    }
    \label{tab:tao}
\end{table}

\paragraph{TAO}
TAO dataset~\cite{tao} annotates 482 classes in total, which are a subset the classes annotated in the LVIS dataset~\cite{lvis}.
It has 400 videos, 216 classes in the training set, 988 videos, 302 classes in the validation set, and 1419 videos, 369 classes in the test set.
The classes in train, validation, and test sets may not overlap.
The videos are annotated at 1 FPS.
The annotated classes in TAO follow a long-tailed distribution, \eg, half of the annotated instances are of class person and a sixth of the objects are of class car, while there are many classes with only few annotated instances.


\subsection{Metrics}
We use several well-established tracking metrics for evaluation.

\paragraph{MOTA}
The Multiple Object Tracking Accuracy (MOTA)~\cite{clearmot} metric computes tracking accuracy in tandem with detection accuracy.
It is defined as,
\begin{equation}
    \texttt{MOTA} = 1 - \frac{\sum_t \left( m_t + f_t + e_t \right)}{\sum_t g_t},
\end{equation}
where $t$ is the timestep, $m_t$ is the number of misses, $f_t$ is the number of false positives, $e_t$ is the number of mismatches, and $g_t$ is the number of objects.
MOTA weighs detection performance more heavily than association performance.
For tracking with multiple classes, we compute MOTA for each class independently then take an average over the number of classes (mMOTA).

\paragraph{IDF1}
The Identification $F_1$ Score (IDF1)~\cite{ristani2016performance} matches ground truth and predictions on the trajectory level and computes a corresponding F1-score. 
It is defined as,
\begin{equation}
    \texttt{IDF1} = \frac{|\texttt{IDTP}|}{|\texttt{IDTP}| + 0.5 |\texttt{IDFN}| + 0.5|\texttt{IDFP}|},
\end{equation}
where \texttt{IDTP}, \texttt{IDFN}, and \texttt{IDFP} are the true positive, false negative, and false positive trajectories.
IDF1 focuses on measuring association performance.
Similar to MOTA, we compute an average over multiple classes for multi-class tracking (mIDF1).

\paragraph{HOTA}
Higher Order Tracking Accuracy (HOTA)~\cite{luiten2021hota} aims to fairly combine the evaluation of detection and association. Therefore, HOTA is composed of two accuracy scores, detection accuracy \texttt{DetA} and association accuracy \texttt{AssA}.
\texttt{DetA} is defined as,
\begin{equation}
    \texttt{DetA} = \frac{|\texttt{TP}|}{|\texttt{TP}| + |\texttt{FN}| + |\texttt{FP}|},
\end{equation}
where \texttt{TP}, \texttt{FN}, and \texttt{FP} are the true positive, false negative, and false positive detections.
Additionally, detection recall \texttt{DetRe} and detection precision \texttt{DetPr} are used.
\texttt{AssA} is defined as,
\begin{equation}
    \texttt{AssA} = \frac{1}{|\texttt{TP}|} \sum_{a\in\texttt{TP}} \frac{|\texttt{TPA}(a)|}{|\texttt{TPA}(a)| + |\texttt{FNA}(a)| + |\texttt{FPA}(a)|},
\end{equation}
where \texttt{TPA}, \texttt{FNA}, and \texttt{FPA} are the true positive, false negative, and false positive associations.
Similarly, association recall \texttt{AssRe} and association precision \texttt{AssPr} are used.
HOTA is computed as a geometric mean of \texttt{DetA} and \texttt{AssA}.

\subsection{Implementation details}
Two-stage object detectors use a Region Proposal Network (RPN) to first generate a set of proposal bounding boxes, \ie, Region of Interests (RoIs).
We use the RoIs from the RPN for similarity learning.
One-stage object detectors do not have a proposal stage and instead perform detection directly on the entire dense grid of bounding box locations.
As our similarity learning protocol requires object regions, we generate them by simply using the dense detection outputs before post-processing.
We follow the same box filtering procedure as the RPN~\cite{frcnn}, where we keep the most confident 1000 boxes then apply Non-maximum Suppression (NMS) with an IoU threshold of 0.7.
We investigate Faster R-CNN~\cite{frcnn} for two-stage detectors and RetinaNet~\cite{lin2017focal} and YOLOX~\cite{ge2021yolox} for one-stage detectors in this work.

We select 128 RoIs from the key frame as training samples, and 256 RoIs from the reference frame with a positive-negative ratio of 1.0 as contrastive targets.
We use IoU-balanced sampling~\cite{pang2019libra} to sample negative RoIs, which better balances the sampling of hard negatives according to their IoU.
We use \emph{4conv-1fc} head with group normalization~\cite{wu2018group} to extract feature embeddings.
The channel number of embedding features is set to 256 by default.
We keep backdrops only from the previous frame.
For association, we associate objects only when they are classified as the same category.

\begin{figure}
    \centering
    \includegraphics[width=\linewidth]{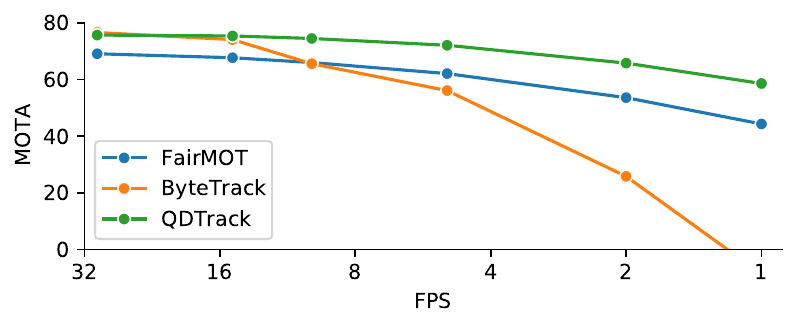}
    \caption{\textbf{Ablation study on video frame rate.} We compare QDTrack to state-of-the-art tracking methods, namely ByteTrack~\cite{zhang2021bytetrack} and FairMOT~\cite{zhang2020fair}, on the MOT17 validation split at different video frame rates.}
    \label{fig:mot17_fps_compare}
\end{figure}

\input{tables/ablation_augmentation}

\input{tables/ablation_detectors}

\input{tables/ablation_inference}

\footnotetext{\url{https://github.com/cheind/py-motmetrics}}

On MOT17 and MOT20, we follow the recent practice of \cite{zhang2021bytetrack, cao2022observation, du2022strongsort} and train QDTrack with the popular YOLOX~\cite{ge2021yolox} detector on the union of CrowdHuman~\cite{shao2018crowdhuman} and the respective MOT benchmark.
On DanceTrack and BDD100K, we again follow~\cite{zhang2021bytetrack} and use the same detector, but we only train on the respective dataset.
For data augmentation, we follow~\cite{ge2021yolox} and utilize MixUp~\cite{zhang2017mixup} and Mosaic augmentations.
For our ablation studies we use Faster-RCNN in combination with ResNet-50 and FPN unless otherwise noted.
%

On Waymo, we use the original scale of the images for training and inference.
We do not use any other data augmentation methods except random horizontal flipping and initialize the base model with ImageNet pre-trained weights for training.
%
%
On TAO, we randomly select a scale between 640 to 800 and resize the shorter side of images during training.
At inference time, the shorter side of the images are resized to 800.
We use an LVIS~\cite{lvis} pre-trained model, consistent with the implementation of~\cite{tao}.
We freeze the detection model and only fine-tune the embedding head to extract instance representations as the annotations in TAO are incomplete.
Since not all objects in the videos are annotated, fine-tuning the detection model will lead to worse performance.
%
We provide a more detailed overview and analysis of our hyper-parameters in the appendix.

\subsection{Comparison to state-of-the-art}
We compare our method to existing literature across five challenging multi-object tracking benchmarks.

\paragraph{MOT}
%
The official benchmark results with private detectors on MOT17 and MOT20 benchmarks are shown in Table~\ref{tab:mot}.
Our method achieves competitive performance on both benchmarks, despite only utilizing appearance cues for association.
Notably, QDTrack obtains a high score of 63.5 HOTA on MOT17 and 60.0 HOTA on MOT20.
Since the MOT benchmarks are captured at a relatively high frame rate and include only limited camera motion, 2D motion based association~\cite{zhang2021bytetrack, du2022strongsort, cao2022observation} works very well in this scenario.
However, this only holds true for the high frame rate scenario.
In Figure~\ref{fig:mot17_fps_compare} we show that when reducing the video frame rate on the MOT17 validation split, the performance of ByteTrack~\cite{zhang2021bytetrack} drops quickly and even completely fails at a frame rate of 1 FPS, while our tracker still achieves 58.6 MOTA at this frame rate.
Furthermore, we show that our tracker also compares favorably to other appearance-based trackers in this regime, namely FairMOT~\cite{zhang2020fair}, which drops to 44.3 MOTA maintaining only 64.1\% of its original performance, while we maintain 77.4\%.

\input{tables/ablation_motion}

\paragraph{DanceTrack}
The results on the benchmark are shown in Table~\ref{tab:comparison_dancetrack}.
Surprisingly, while DanceTrack was specifically designed to provide a platform to develop MOT algorithms that rely less on visual appearance and more on motion analysis, we find that our appearance based tracker performs very well on this dataset, reaching a HOTA score only marginally behind the state-of-the-art method OC-SORT~\cite{cao2022observation} ($-0.9$ HOTA).  We achieve this score without any bells and whistles, naively applying the same configuration as in our MOT17 experiments to train on the DanceTrack dataset, following \cite{zhang2021bytetrack}. This reinforces our argument that one can in fact build a robust tracking algorithm by relying on our quasi-dense instance similarity.

\paragraph{BDD100K}
The main results on the BDD100K tracking validation and testing sets are in Table~\ref{tab:bdd}.
On the validation set, QDTrack with YOLOX-X achieves 42.1 mMOTA and 54.3 mIDF1, which are the second-best results behind ByteTrack~\cite{zhang2021bytetrack}.
Still, QDTrack achieves much better results in IDF1 (73.3 vs. 70.4).
On the test set, QDTrack with YOLOX-X achieves a high score of 42.4 mMOTA, 55.6 mIDF1, and 73.9 IDF1, outperforming all other methods by a significant margin.
In particular, QDTrack outperforms
ByteTrack by 2.3 mMOTA and 2.6 IDF1.
QDTrack with Faster R-CNN also achieves a competitive score of 38.7 mMOTA, 54.1 mIDF1, and 74.0 IDF1, outperforming other methods using the same detector.
TETer~\cite{trackeverything} is an extension of QDTrack that employs a new association strategy designed for improving long-tailed object tracking.
These results demonstrate that QDTrack can perform well even on a more challenging large-scale benchmark with a simple framework.

\paragraph{Waymo}
Table~\ref{tab:waymo} shows our main results on Waymo open dataset.
We report the results on the validation set following the setup of RetinaTrack~\cite{retinatrack}, which only conduct experiments on the vehicle class.
We also report the overall performance for future comparison.
We report the results on the test set via official rules.
Our method outperforms all baselines on both validation set and test set.
We obtain 44.0 MOTA and 56.8 IDF1 on the validation set and 49.4 MOTA/L1 and 43.9 MOTA/L2 on the test set.
The performance of vehicle on the validation set is 10.7, 13.0, and 17.4 points higher than RetinaTrack~\cite{retinatrack}, Tracktor++~\cite{tracktor, retinatrack}, and IoU baseline~\cite{retinatrack}, respectively.
Our model with ResNet-101 and deformable convolution (DCN)~\cite{deformable} has state-of-the-art performance on the test benchmark, which is on par with the champion of Waymo 2020 2D Tracking Challenge (HorizonMOT) despite only using a simple single model.

\paragraph{TAO}
The results for TAO are shown in Table~\ref{tab:tao}.
We obtain 16.1 AP50 on the validation set and 12.4 AP50 on the test set.
The results are 2.9 points and 2.2 points higher than TAO's baseline.
Although we only boost the overall performance by 2 to 3 points, we outperform the baseline by a large margin on frequent classes, \ie, 38.6 points vs. 18.5 points on person.
This improvement is not well represented in the standard evaluation metrics of TAO, since it averages per-class scores across hundreds of classes.
GTR~\cite{zhou2022global} and AOA~\cite{du2021aoa} are recent methods proposed to tackle long-tail multi-object tracking.
Although they outperform our method, GTR is an offline method and AOA utilizes separate ReID networks trained on additional data.

\begin{figure*}[t]
    \centering
    \includegraphics[width=\linewidth]{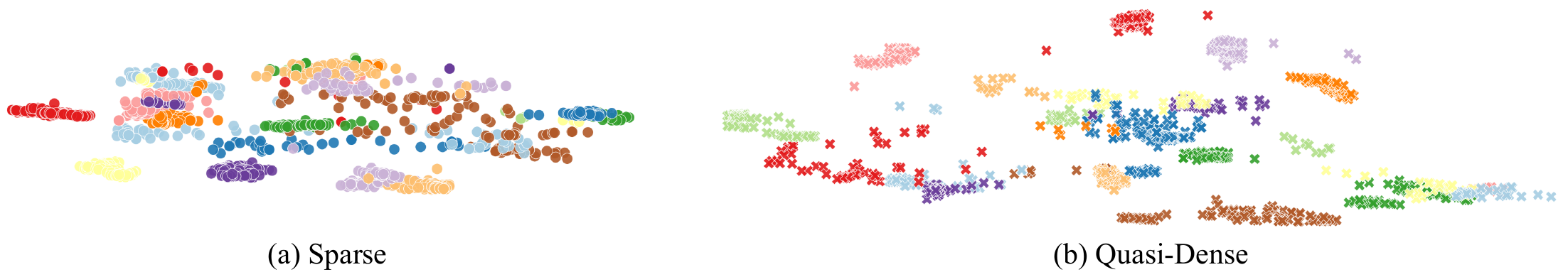}
    \caption{\textbf{Instance embedding space visualization}. We visualize the instance embedding space learned via (a) sparse matching and (b) quasi-dense matching using t-SNE. We show ground truth embedding identities as color and plot embedding vectors sampled from a sequence in the BDD100K tracking validation set.}
    \label{fig:embeds}
    \vspace{-4mm}
\end{figure*}

\subsection{Learning to track from static images}
\label{subsec:selfsup}
Since our quasi-dense instance similarity learning is agnostic to how the image pair is generated during training, we investigate how we can leverage static images where no association annotations are available. Inspired by recent literature in self-supervised representation learning~\cite{he2020momentum, chen2020simple}, we experiment with different data augmentations on static images to learn discriminative instance representations from static input. 
In particular, for a given training sample in a detection dataset, we generate two distorted images via data augmentation techniques. 
We find that random horizontal flip (HF), multi-scale resize and crop (MS), color jittering (Color), and MixUp / Mosaic augmentations are the most suitable for our use-case.
If the augmentation parameters are not shared across the key and reference view, we denote it with `NC' (non-consistent).
We only use MixUp / Mosaic with consistent parameters in order to compose the same images between key and reference views and thus be able to match objects across them. 
To train our models, we utilize the detection (image) and tracking (video) splits of BDD100K.
Note that the tracking split contains much more data, thus influencing the detection performance.

The results of our experiments are shown in Table~\ref{tab:ablation_aug}.
We use Faster-RCNN~\cite{frcnn} as the detector with ResNet-50~\cite{resnet} and FPN~\cite{fpn} as the base model and evaluate the tracking performance on the BDD100K tracking benchmark~\cite{bdd100k}.
We observe that when we only apply consistent HF, the tracking performance is far behind the version trained with full tracking supervision.
By adding in non-consistent augmentations and MixUp / Mosaic, we can narrow this gap and achieve comparable accuracy to the fully supervised model.
In particular, we exceed the mMOTA of the fully supervised baseline trained without augmentations besides HF when training on the same amount of training data by a significant margin.
This clearly shows that not only detection, but also association benefits greatly from the data augmentation, and that with proper data augmentation, our similarity learning scheme can track objects effectively while trained on static images alone. If we use the same amount of training data, we indeed rival the performance of the best supervised model, shown by the small gap in mMOTA ($-0.1$ points).

In addition, we observe that the data augmentation scheme can also benefit the supervised models, reaching a much higher score than in our initial work~\cite{qdtrack} without changing the network architecture ($+1.2$ points in mMOTA, $+2.1$ points in mIDF1). The increase in mIDF1 highlights the benefit of data augmentation to the robustness of instance similarity learning.

\subsection{Ablation studies}
We conduct ablation studies on the validation set of BDD100K~\cite{bdd100k}, where we investigate the importance of the major model components for training and inference procedures.

\paragraph{Different object detectors, feature extractors, and training schedules}
We combine our method with different object detectors and feature extractors to verify the flexibility of our instance similarity learning scheme.
In Table~\ref{tab:ablation_det}, we show the tracking performance of our method with ResNet-50, ResNet-101~\cite{resnet}, as well as the modified CSPNet~\cite{wang2020cspnet} on the tracking validation set of BDD100K.
We combine those feature extractors with a Faster-RCNN~\cite{frcnn} detector and observe that ResNet-101 achieves the best performance with 66.2 MOTA, 73.1 IDF1, and 35.3 AP.

In addition, we apply our method on two more base object detection models, namely RetinaNet~\cite{lin2017focal} and YOLOX~\cite{ge2021yolox}.
Both methods produce reasonable results, and the YOLOX model achieves the best overall scores with 68.2 MOTA, 73.3 IDF1, and 38.9 AP.
It shows that our method can work independent of feature extractor or base detection model.
Finally, we also experiment with different training schedules.
We investigate the effect of longer training, increasing the epochs from 12 (1x schedule) to 24 (2x schedule) and 25.
Note that we use the extensive data augmentation techniques presented in section~\ref{subsec:selfsup} in this ablation study to counteract overfitting when training with longer schedules.
We find that increasing the number of epochs does not help the smaller ResNet models, but is beneficial for training very large models like YOLOX-X.

\paragraph{Importance of quasi-dense matching}
The results are presented in the top sub-table of Table~\ref{tab:ablation_inf}.
We use a Faster R-CNN detector with ResNet-50 base model.
MOTA and IDF1 are calculated over all instances without considering categories as overall evaluations.
We use cosine distance to calculate the similarity scores during the inference procedure.
Compared to learning with sparse ground truths, quasi-dense tracking improves the overall IDF1 by 4.8 points (63.0\% to 67.8\%).
The significant improvement on IDF1 indicates quasi-dense tracking greatly improves the feature embeddings and enables more accurate associations.

We then analyze the improvements in detail.
In the table, we can observe that when we match each training sample to more negative samples and train the feature space with Eq.~\eqref{eqa:npair},
the IDF1 is significantly improved by 3.4 points.
This improvement contributes 70\% to the total improved 4.8 points IDF1.
This experiment shows that more contrastive targets, even most of them are negative samples, can improve the feature learning process.
The multiple-positive contrastive learning following Equation~\eqref{eqa:multipos} further improves the IDF1 by 1 point (66.8\% to 67.8\%).

\paragraph{Importance of bi-softmax}
We investigate how different inference strategies influence the performance.
As shown in the bottom of Table~\ref{tab:ablation_inf}, replacing cosine similarity by bi-softmax improves overall IDF1 by 2.2 points and the IDF1 of pedestrian by 4.5 points.
This experiment also shows that the one-to-one constraint further strengthens the estimated similarity.

\input{figures/limitations/limitations}

\input{tables/bdd100k_mots}

\paragraph{Importance of matching candidates}
Duplicate removal and backdrops improve IDF1 by 1.5 points.
Overall, our training and inference strategies  improve the IDF1 by 8.5 points (63.0\% to 71.5\%).
The total number of ID switches is decreased by 30\%.
Especially, the MOTA and IDF1 of pedestrian are improved by 9.1 points and 10.5 points respectively, which further demonstrate the power of quasi-dense contrastive learning.

\paragraph{Combinations with motion and location}
Finally, we try to add location and motion priors to understand whether they are still helpful when we have good feature embeddings for measuring similarity.
These experiments follow the procedures in Tracktor~\cite{tracktor} and use the same detector for fair comparisons.
As shown in Table~\ref{tab:ablation_mot}, without appearance features, the tracking performance is consistently improved with the introduction of additional information.
However, these cues barely enhance the performance of our approach.
Our method yields the best results when only using appearance embeddings.
The results indicate that our instance feature embeddings are sufficient for multiple object tracking with the effective quasi-dense matching, which greatly simplifies the inference pipeline.

\paragraph{Inference speed}
To understand the runtime efficiency, we profile our method on a single NVIDIA RTX 3090 graphics card.
Because it only adds a lightweight embedding head to the detector, our method only causes marginal overhead in inference speed. With an input size of $1296 \times 720$ and a Faster R-CNN detector with ResNet-50 base model on BDD100K, the inference time is 61 ms, equating to 16.3 FPS. However, the embedding extractor consumes 3 ms, representing only 5\% of the total runtime.

\subsection{Embedding visualizations}
We use t-SNE to visualize the embeddings trained with sparse matching and our quasi-dense matching and show them in Figure~\ref{fig:embeds}.
The instances are selected from a video in BDD100K tracking validation set.
The same instance is shown with the same color.
We observe that it is easier to separate objects in the feature space of quasi-dense matching.

\subsection{Segmentation tracking}
Owing to the simplicity of our method, we can extend it to instance segmentation tracking in a straightforward manner.
To do so, we simply add a Mask R-CNN~\cite{maskrcnn} mask prediction head to the existing network architecture and use a pre-trained QDTrack model trained on the BDD100K tracking set to fine-tune the mask head on MOTS data. In particular, BDD100K provides a subset for the segmentation tracking task. There are 154 videos in the training set, 32 videos in the validation set, and 37 videos in the test set.
Table~\ref{tab:bddtrack} shows the results on the BDD100K segmentation tracking task compared to other methods.
QDTrack achieves 25.6 mMOTSA and 45.2 mIDF1.
PCAN~\cite{ke2021prototypical} is an extension of QDTrack that utilizes a prototypical appearance module to further improve segmentation.
We observe that QDTrack based models achieve much better performance than previous methods.

\subsection{Limitations}

While our method gains in simplicity and generality by solely relying on instance similarity learning, we also identify certain challenges that arise with this paradigm.
In particular, we observe that our model struggles with rapid changes in object appearance, \eg, through partial occlusion.
Also, since our model relies on discrete class labels to aid the matching process, we observe that classification errors can lead to truncated object tracks.
These cases are illustrated in Figure~\ref{fig:limitations}.
In addition, inaccurate object localization can lead to difficulties in association when regions within a bounding box cover background and/or other objects, thus impeding accurate instance embedding extraction.
For more detailed failure case and oracle analysis, please refer to the appendix.

%% file: tables/comparison_mot.tex
\setlength{\tabcolsep}{6pt}
\begin{table*}[t]
    \caption{
        \textbf{Comparison to state-of-the-art on MOT Challenge benchmarks.}
        We benchmark our method against existing works on both MOT17 and MOT20 test sets with private detections.
        Datasets include CrowdHuman (CH)~\cite{shao2018crowdhuman}, CityPersons (CP)~\cite{zhang2017citypersons}, and ETHZ~\cite{ess2008ethz}.
        $\dagger$ indicates using COCO pre-trained weights.
        $\uparrow$ means higher is better.
    }
    \centering
    \resizebox{\linewidth}{!}{
    \small
    \setlength\tabcolsep{2.4mm}
    \begin{tabular}{lcccccccccccc}
        \toprule
        Method & Detector & Base model & Datasets & MOTA $\uparrow$ & IDF1 $\uparrow$ & HOTA $\uparrow$ & AssA $\uparrow$ & DetA $\uparrow$ & AssRe $\uparrow$ & AssPr $\uparrow$ & DetRe $\uparrow$ & DetPr $\uparrow$ \\
        \midrule
        
        \textbf{MOT17} \\
        \midrule
        CenterTrack~\cite{centertrack} & CenterNet~\cite{zhou2019objects} & DLA-34 & MOT, CH & 67.8 & 64.7 & 52.2 & 51.0 & 53.8 & 56.6 & 73.0 & 57.5 & 76.9 \\
        FairMOT~\cite{zhang2020fair}~$\dagger$ & CenterNet~\cite{zhou2019objects} & DLA-34 & MOT, CH, CP, ETHZ & 73.7 & 72.3 & 59.3 & 58.0 & 60.9 & 63.6 & 76.3 & 66.0 & 78.5 \\
        ReMOT~\cite{YANG2021104091} & - & - & - & 77.0 & 72.0 & 59.7 & 57.1 & 62.8 & 61.7 & 78.0 & 68.8 & 77.1  \\
        OC-SORT~\cite{cao2022observation}~$\dagger$ & YOLOX-X & Modified CSP & MOT, CH, CP, ETHZ & 78.0 & 77.5 & 63.2 & 63.4 & 63.2 & 67.5 & 80.8 & 67.2 & 80.3 \\
        MAA~\cite{maatrack} & CrowdDet~\cite{chu2020detection} & R50-FPN & MOT, CH & 79.4 & 75.9 & 62.0 & 60.2 & 64.2 & 67.3 & 74.0 & 70.9 & 76.4\\
        StrongSORT~\cite{du2022strongsort}~$\dagger$ & YOLOX-X & Modified CSP & MOT, CH, CP, ETHZ & 79.6 & 79.5 & 64.4 & 64.4 & 64.6 & 71.0 & 78.7 & 70.2 & 78.3 \\
        ByteTrack~\cite{zhang2021bytetrack}~$\dagger$ & YOLOX-X & Modified CSP & MOT, CH, CP, ETHZ & 80.3 & 77.3 & 63.1 & 62.0 & 64.5 & 68.2 & 76.0 & 70.1 & 78.1 \\ 
         \midrule
        \multirow{2}{*}{QDTrack (Ours)~$\dagger$} & FRCNN & R50-FPN & MOT, CH & 77.2 & 72.2 & 58.8 & 56.2 & 61.8 & 62.6 & 74.1 & 66.6 & 78.1 \\
         & YOLOX-X & Modified CSP & MOT, CH & 78.7 & 77.5 & 63.5 & 62.6 & 64.5 & 69.3 & 76.2 & 71.0 & 77.7 \\
        \midrule
        \textbf{MOT20} \\
        \midrule
        SGT~\cite{hyun2022detection}~$\dagger$ & CenterNet~\cite{zhou2019objects} & DLA-34 & MOT, CH & 72.8 & 70.5 & 56.9 & 55.3 & 58.8 & 60.3 & 75.4 & 63.8 & 76.8 \\
        StrongSORT~\cite{du2022strongsort}~$\dagger$ & YOLOX-X & Modified CSP & MOT, CH & 73.8 & 77.0 & 62.6 & 64.0 & 61.3 & 69.6 & 80.0 & 65.3 & 81.2 \\
        MAA~\cite{maatrack} & CrowdDet~\cite{chu2020detection} & R50-FPN & MOT, CH & 73.9 & 71.2 & 57.3 & 55.1 & 59.7 & 61.1 & 72.1 & 64.8 & 77.4 \\
        OC-SORT~\cite{cao2022observation}~$\dagger$ & YOLOX-X & Modified CSP & MOT, CH & 75.7 & 76.3 & 62.4 & 62.5 & 62.4 & 67.4 & 79.6 & 66.9 & 80.4 \\
        ReMOT~\cite{YANG2021104091} & - & - & - & 77.4 & 73.1 & 61.2 & 58.7 & 63.9 & 63.1 & 79.5 & 69.8 & 78.6	 \\
        ByteTrack~\cite{zhang2021bytetrack}~$\dagger$ & YOLOX-X & Modified CSP & MOT, CH & 77.8 & 75.2 & 61.3 & 59.6 & 63.4 & 66.2 & 74.6 & 69.1 & 78.4 \\
        \midrule
        QDTrack (Ours)~$\dagger$ & YOLOX-X & Modified CSP & MOT, CH & 74.7 & 73.8 & 60.0 & 58.9 & 61.4 & 65.7 & 74.8 & 66.4 & 79.1 \\
        \bottomrule
    \end{tabular}
    }
    \label{tab:mot}
\end{table*}
\setlength{\tabcolsep}{6pt}

%% file: tables/comparison_dancetrack.tex
\begin{table}[t]
    \caption{\textbf{Comparison to state-of-the-art on DanceTrack.} We compare our method to existing methods on the challenging DanceTrack test set. We use YOLOX-X~\cite{ge2021yolox} as our detector.}
    \centering
    \resizebox{\linewidth}{!}{
        \begin{tabular}{c|ccccc}
            \toprule
            Method & HOTA $\uparrow$ & DetA $\uparrow$ & AssA $\uparrow$ & MOTA $\uparrow$ & IDF1 $\uparrow$ \\
            \midrule
FairMOT~\cite{zhang2020fair} & 39.7 & 66.7 & 23.8 & 82.2 & 40.8 \\
CenterTrack~\cite{centertrack} & 41.8 & 78.1 & 22.6 & 86.8 & 35.7\\
TraDes~\cite{wu2021track} & 43.3 & 74.5 & 25.4 & 86.2 & 41.2\\
TransTrack~\cite{transtrack} & 45.5 & 75.9 & 27.5 & 88.4 & 45.2\\
ByteTrack~\cite{zhang2021bytetrack} & 47.7 & 71.0 & 32.1 & 89.6 & 53.9 \\
GTR~\cite{zhou2022global} & 48.0 & 72.5 & 31.9 & 84.7 &  50.3\\
MOTR~\cite{zeng2021motr} & 54.2 & 73.5 & 40.2	& 79.7 & 51.5 \\
OC-SORT~\cite{cao2022observation} & 55.1 & 80.3 & 38.3 &  92.0 & 54.6 \\
\midrule
QDTrack (Ours) & 54.2 & 80.1 & 36.8 & 87.7 & 50.4\\
            \bottomrule
        \end{tabular}
    }
    \label{tab:comparison_dancetrack}
\end{table}

%% file: tables/comparison_bdd100k.tex
\setlength{\tabcolsep}{4pt}
\begin{table*}[t]
    \caption{\textbf{Comparison to state-of-the-art on BDD100K.} We report results and compare with existing works on the BDD100K tracking validation and test set.
    $\dagger$ indicates using COCO pre-trained weights.
    }
    \centering
    \resizebox{\linewidth}{!}{
        \begin{tabular}{clccccccccccc}
            \toprule
            Split                  & Method & Detector & Base model & mMOTA $\uparrow$ & mIDF1 $\uparrow$ & MOTA $\uparrow$ & IDF1 $\uparrow$ & FN $\downarrow$ & FP $\downarrow$ & ID Sw. $\downarrow$ & MT $\uparrow$  & ML $\downarrow$ \\ 
            \midrule
            \multirow{5}{*}{val}  & Yu~\etal~\cite{bdd100k} & FRCNN & DLA-34 & 25.9             & 44.5             & 56.9            & 66.8            & 122406          & 52372           & 8315                & 8396           & 3795            \\ 
             & DeepSORT~\cite{deepsort} & FRCNN & R50-FPN & 35.2 & 49.3 & - & - & - & - & - & - & - \\
             & TETer~\cite{trackeverything} & FRCNN & R50-FPN & 39.1 & 53.3 & - & - & - & - & - & - & - \\
             & ByteTrack~\cite{zhang2021bytetrack}~$\dagger$ & YOLOX-X & Modified CSP & 45.5 & 54.8 & 69.1 & 70.4 & 92805 & 34998 & 9140 & 9626 & 3005 \\ 
            \cmidrule{2-13}
              & QDTrack (Ours) & FRCNN & R50-FPN & 37.7    & 52.9    & 65.7   & 72.7   & 104861 & 41355  & 5640       & 9649  & 2874   \\
              & QDTrack (Ours)~$\dagger$ & YOLOX-X & Modified CSP & 42.1    & 54.3    & 68.2   & 73.3   & 83395 & 48798  & 8478       & 10925  & 2272   \\ 
            \midrule
            \multirow{7}{*}{test}  & Yu~\etal~\cite{bdd100k} & FRCNN & DLA-34 & 26.3             & 44.7             & 58.3            & 68.2            & 213220          & 100230          & 14674               & 16299          & 6017            \\ 
              & DeepBlueAI & - & -              & 31.6             & 38.7             & 56.9            & 56.0            & 292063          & 35401  & 25186               & 10296          & 12266           \\ 
              & madamada & - & -                & 33.6             & 43.0             & 59.8            & 55.7            & 209339          & 76612           & 42901               & 16774          & 5004   \\ 
             & DeepSORT~\cite{deepsort} & FRCNN & R50-FPN & 34.0 & 50.2 & - & - & - & - & - & - & - \\
             & TETer~\cite{trackeverything} & FRCNN & R50-FPN & 37.4 & 53.3 & - & - & - & - & - & - & - \\
             & ByteTrack~\cite{zhang2021bytetrack}~$\dagger$ & YOLOX-X & Modified CSP & 40.1 & 55.8 & 69.9 & 71.3 & 169073 & 63869 & 15466 & 18057 & 5107 \\ 
            \cmidrule{2-13}
              & QDTrack (Ours) & FRCNN & R50-FPN                    & 38.7    & 54.1    & 66.5   & 74.0   & 185773 & 78068           & 10098      & 18167 & 4635            \\
              & QDTrack (Ours)~$\dagger$ & YOLOX-X & Modified CSP                    & 42.4    & 55.6    & 68.4   & 73.9   & 154797 & 89376           & 14282      & 19852 & 3924            \\ 
            \bottomrule
        \end{tabular}
    }
    \label{tab:bdd}
\end{table*}
\setlength{\tabcolsep}{6pt}

%% file: tables/comparison_waymo.tex
\begin{table*}[t]
    \caption{\textbf{Comparison to state-of-the-art on Waymo.} We show results of our method compared with existing methods on the Waymo Open tracking validation set using py-motmetrics library (top) \protect \footnotemark and test set using official evaluation (bottom). We use Faster R-CNN~\cite{frcnn} as our detector.
        *~indicates methods using undisclosed detectors.
    }
    \centering
    \resizebox{\linewidth}{!}{
        \begin{tabular}{clccccccccccc}
            \toprule
            Split  & Method & Category & MOTA $\uparrow$    & IDF1 $\uparrow$    & FN $\downarrow$      & FP $\downarrow$      & ID Sw. $\downarrow$ & MT $\uparrow$      & ML $\downarrow$      & mAP $\uparrow$       \\
            \midrule
            \multirow{5}{*}{val} & IoU baseline~\cite{retinatrack}        & Vehicle  & 38.3              & -                  & -                    & -                    & -                   & -                  & -                    & 45.8                \\
             & Tracktor++~\cite{tracktor,retinatrack}  & Vehicle  & 42.6              & -                  & -                    & -                    & -                   & -                  & -                    & 42.4                \\
             & RetinaTrack~\cite{retinatrack}    & Vehicle  & 44.9              & -                  & -                    & -                    & -                   & -                  & -                    & 45.7                \\
            \cmidrule{2-11}
             & \multirow{2}{*}{QDTrack (Ours)}    & Vehicle  & \textbf{55.6}               & 66.2               & 514548               & 214998               & 24309               & 17595              & 5559                 & \textbf{49.5}                 \\
            &   & All      & 44.0               & 56.8               & 674064               & 264886               & 30712               & 21410              & 7510                 & 40.1                 \\
            \bottomrule
            \toprule
            Method                                 & Split & Category & MOTA/L1 $\uparrow$ & FP/L1 $\downarrow$ & MisM/L1 $\downarrow$ & Miss/L1 $\downarrow$ & MOTA/L2 $\uparrow$  & FP/L2 $\downarrow$ & MisM/L2 $\downarrow$ & Miss/L2 $\downarrow$ \\
            \midrule
            \multirow{5}{*}{test} & Tracktor~\cite{kim2018, waymo}         & Vehicle  & 34.8              & 10.6              & 14.9                & 39.7                & 28.3               & 8.6               & 12.1                & 51.0                \\
             & CascadeRCNN-SORTv2* & All      & 50.2              & 7.8               & 2.7                 & 39.3                & 44.2               & \textbf{6.9}      & 2.4                 & 46.5                \\
             & HorizonMOT*  & All      & 51.0              & 7.5               & 2.4                 & \textbf{39.0}       & \textbf{45.1}      & 7.1               & 2.3                 & \textbf{45.5}       \\
            \cmidrule{2-11}
             & Ours (ResNet-50)  & All      & 49.4              & \textbf{7.4}      & 1.5                 & 41.7                & 43.9               & 7.1               & \textbf{1.3}        & 48.2                \\
             & Ours (ResNet-101 + DCN)  & All      & \textbf{51.2}     & 7.6               & \textbf{1.5}        & 39.7                & 45.1               & 7.2               & \textbf{1.3}        & 46.4                \\
            \bottomrule
        \end{tabular}
    }
    \label{tab:waymo}
\end{table*}

%% file: tables/ablation_augmentation.tex
\begin{table*}[t]
    \caption{\textbf{Learning to track from static images.} We  apply quasi-dense instance similarity learning on static images without tracking annotations. We use different data augmentation strategies to distort the image pair: horizontal flip (HF), multi-scale resize and crop (MS), color jittering (Color), and MixUp / Mosaic. We denote non-consistent augmentation parameters between key and reference images as `NC'. We evaluate performance on the BDD100K tracking validation set and compare to a model trained with video input and tracking annotations.}
    \centering
    \begin{tabular}{cc|ccccc|ccccc}
        \toprule
        \multirow{2}{*}{Input} & \multirow{2}{*}{Supervision}  & \multicolumn{5}{c}{Augmentations}  & \multicolumn{5}{|c}{BDD100K} \\
        ~ & ~ & HF & HF-NC &MS-NC & Color-NC & MixUp / Mosaic  & mMOTA $\uparrow$ & mIDF1 $\uparrow$ & MOTA $\uparrow$ & IDF1 $\uparrow$ & AP $\uparrow$ \\
        \midrule
        \multirow{5}{*}{image} & \multirow{5}{*}{detection} & \checkmark & - & - & - & - & 32.0 & 43.4 & 54.7 & 58.4 & 29.8 \\
         &  & \checkmark & - & \checkmark & - & - & 35.2 & 47.3 & 62.1 & 66.8 & 32.8 \\
         &  & - & \checkmark & \checkmark & - & - & 35.6 & 48.1 & 62.4 & 67.8 & 32.6 \\
         &  & - & \checkmark & \checkmark & \checkmark & - & 36.3 & 48.0 & 62.2 & 67.8 & 32.8 \\
         &  & - & \checkmark & \checkmark & \checkmark & \checkmark & 35.2 & 47.7 & 62.4 & 67.5 & 33.0 \\
        \midrule
        \multirow{5}{*}{image / video} & \multirow{5}{*}{detection} & \checkmark & - & - & - & - & 32.4 & 45.4 & 56.8 & 60.6 & 32.0 \\
         &  & \checkmark & - & \checkmark & - & - & 36.7 & 49.9 & 61.6 & 66.9 & 33.7 \\
         &  & - & \checkmark & \checkmark & - & - & 35.1 & 50.4 & 61.8 & 68.0 & 33.5 \\
         &  & - & \checkmark & \checkmark & \checkmark & - & 36.1 & 50.4 & 62.1 & 68.3 & 34.2 \\
         &  & - & \checkmark & \checkmark & \checkmark & \checkmark & \textbf{37.7} & 51.1 & 62.9 & 68.2 & \textbf{35.2} \\
        \midrule
        \multirow{3}{*}{image / video} & \multirow{3}{*}{tracking} & \checkmark & - & - & - & - & 36.6 & 50.8 & 63.5 & 71.5 & 33.0 \\
         &  & - & \checkmark & \checkmark & \checkmark & - & 36.9 & 52.1 & 64.1 & 71.7 & 34.1 \\
         &  & - & \checkmark & \checkmark & \checkmark & \checkmark & \textbf{37.7} & \textbf{52.9} & \textbf{65.7} & \textbf{72.7} & 35.0 \\
        \bottomrule
   \end{tabular}
    \label{tab:ablation_aug}
\end{table*}

%% file: tables/ablation_detectors.tex
\setlength{\tabcolsep}{6pt}
\begin{table}[t]
    \caption{\textbf{Ablation study on different training schedules, feature extractors, and detectors.} We train multiple QDTrack models with different detectors, feature extractors and training schedules and measure tracking performance on the BDD100K tracking validation set.}
    \centering
    \resizebox{\linewidth}{!}{
        \begin{tabular}{cccccc}
            \toprule
            Detector & Base model & Schedule & MOTA $\uparrow$ & IDF1 $\uparrow$ & AP $\uparrow$  \\
            \midrule
            \multirow{4}{*}{FRCNN} & R50-FPN & 1x & 65.7 & 72.7 & 35.0  \\
             & R50-FPN & 2x & 65.6 & 73.1 & 35.1 \\
             & R101-FPN & 1x & 66.2 & 73.1 & 35.3  \\
             & R101-FPN & 2x & 65.6 & 72.7 & 34.6  \\
            \midrule
            RetinaNet & R50-FPN & 1x & 60.8  & 69.5 & 32.1 \\
            YOLOX-X & Modified CSP & 25 epochs & \textbf{68.2} & \textbf{73.3} & \textbf{38.9} \\
            \bottomrule
        \end{tabular}
    }
    \label{tab:ablation_det}
\end{table}
\setlength{\tabcolsep}{6pt}

%% file: tables/ablation_inference.tex
\begin{table*}[t]
    \caption{\textbf{Ablation study on quasi-dense matching and inference strategy.} We investigate the contribution of various components on the BDD100K tracking validation set. All models are comparable on detection performance. D. R. means duplicate removal. (P) means results of the class ``pedestrian''.}
    \centering
    \resizebox{\linewidth}{!}{
        \begin{tabular}{ccccccccccc}
            \toprule
            \multicolumn{2}{c}{Quasi-Dense} & \multirow{2}{*}{Metric} & \multicolumn{2}{c}{Matching candidates} & \multirow{2}{*}{MOTA $\uparrow$} & \multirow{2}{*}{IDF1 $\uparrow$} & \multirow{2}{*}{mMOTA $\uparrow$} & \multirow{2}{*}{mIDF1 $\uparrow$} & \multirow{2}{*}{MOTA(P) $\uparrow$} & \multirow{2}{*}{IDF1(P) $\uparrow$}                                  \\
            one-positive                    & multi-positive          &                                         & ~D. R.                           & Backdrops                                                                                                                                                                                                             \\
            \midrule
            -                               & -                       & \emph{cosine}                           & -                                & -                                & 60.4                              & 63.0                              & 34.0                                & 47.9                                & 37.6          & 49.7           \\
            \checkmark                      & -                       & \emph{cosine}                           & -                                & -                                & 61.5                              & 66.8                              & 35.5                                & 50.0                                & 40.5          & 52.7           \\
            -                               & \checkmark              & \emph{cosine}                           & -                                & -                                & 62.5                              & 67.8                              & 36.2                                & 50.0                                & 44.0          & 54.3           \\
            -                               & \checkmark              & \emph{bi-softmax}                       & -                                & -                                & 62.9                              & 70.0                              & 35.4                                & 48.5                                & 45.5          & 58.8           \\
            -                               & \checkmark              & \emph{bi-softmax}                       & \checkmark                       & -                                & 63.2                              & 70.1                              & 36.4                                & 50.4                                & 45.5          & 58.3           \\
            -                               & \checkmark              & \emph{bi-softmax}                       & \checkmark                       & \checkmark                       & \textbf{63.5}                     & \textbf{71.5}                     & \textbf{36.6}                       & \textbf{50.8}                       & \textbf{46.7} & \textbf{60.2}  \\
            \midrule
                                            &                         &                                         &                                  &                                  & \textbf{+3.1}                     & \textbf{+8.5}                     & \textbf{+2.6}                       & \textbf{+2.9}                       & \textbf{+9.1} & \textbf{+10.5} \\
            \bottomrule
        \end{tabular}
    }
    \label{tab:ablation_inf}
\end{table*}

%% file: tables/ablation_motion.tex
\begin{table}[t]
    \caption{\textbf{Ablation study on location and motion cues.} We investigate if our method benefits from using a range of motion priors on the BDD100K tracking validation set. We integrate bounding box IoU, a simple linear motion model, and displacement regression into the association procedure.
    }
    \centering
    \resizebox{\linewidth}{!}{
        \begin{tabular}{cccccccccccc}
            \toprule
            Appearance & IoU        & Motion     & Regression & mMOTA $\uparrow$ & mIDF1 $\uparrow$ \\
            \midrule
            -          & \checkmark & -          & -          & 26.3             & 36.0             \\
            -          & \checkmark & \checkmark & -          & 27.7             & 38.5             \\
            -          & \checkmark & -          & \checkmark & 28.6             & 39.3             \\
            \midrule
            \checkmark & -          & -          & -          & \textbf{36.6}    & \textbf{50.8}    \\
            \checkmark & \checkmark & -          & -          & 36.3             & 49.8             \\
            \checkmark & \checkmark & \checkmark & -          & 36.4             & 49.9             \\
            \checkmark & \checkmark & -          & \checkmark & 36.4             & 50.1             \\
            \bottomrule
        \end{tabular}
    }
    \label{tab:ablation_mot}
\end{table}

%% file: figures/limitations/limitations.tex
\begin{figure*}[t]
\centering
\setlength{\tabcolsep}{1pt}
\resizebox{\linewidth}{!}{
\begin{tabular}{cccc}
$t$ & $t+1$ & $t+2$ & $t+3$ \\
\midrule
\includegraphics[width=0.25\linewidth]{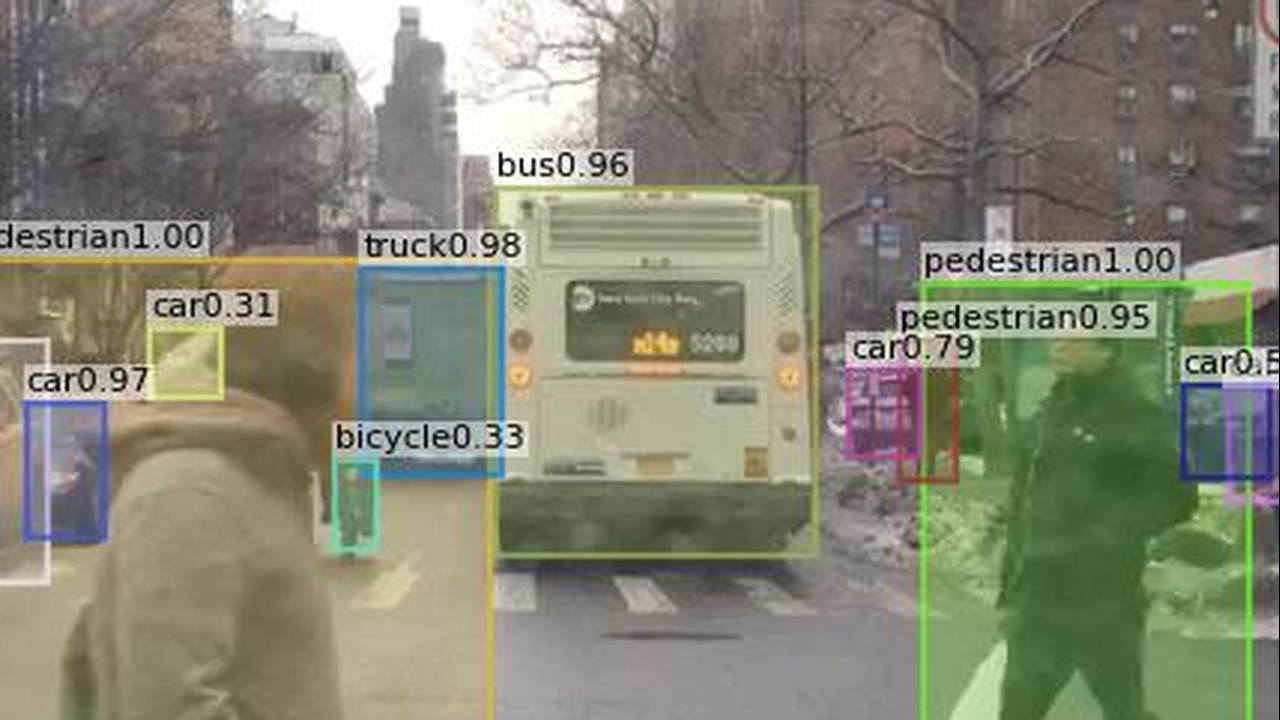} &
\includegraphics[width=0.25\linewidth]{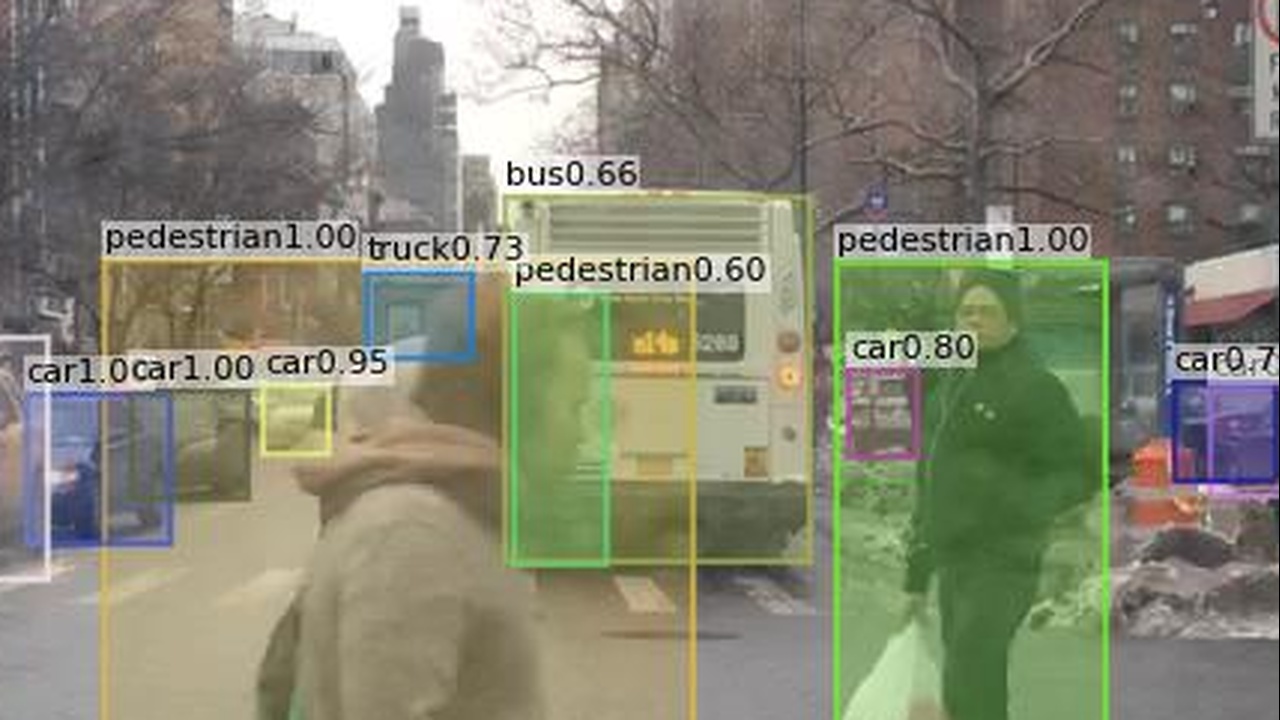} &
\includegraphics[width=0.25\linewidth]{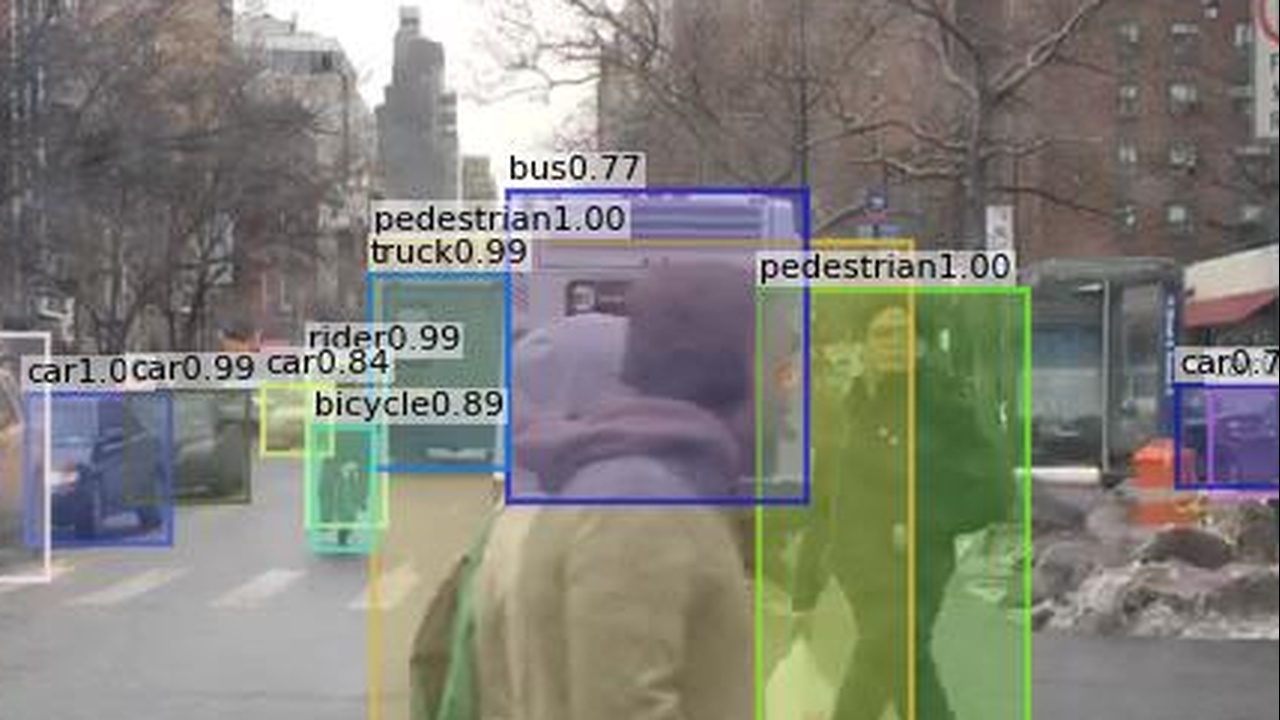} &
\includegraphics[width=0.25\linewidth]{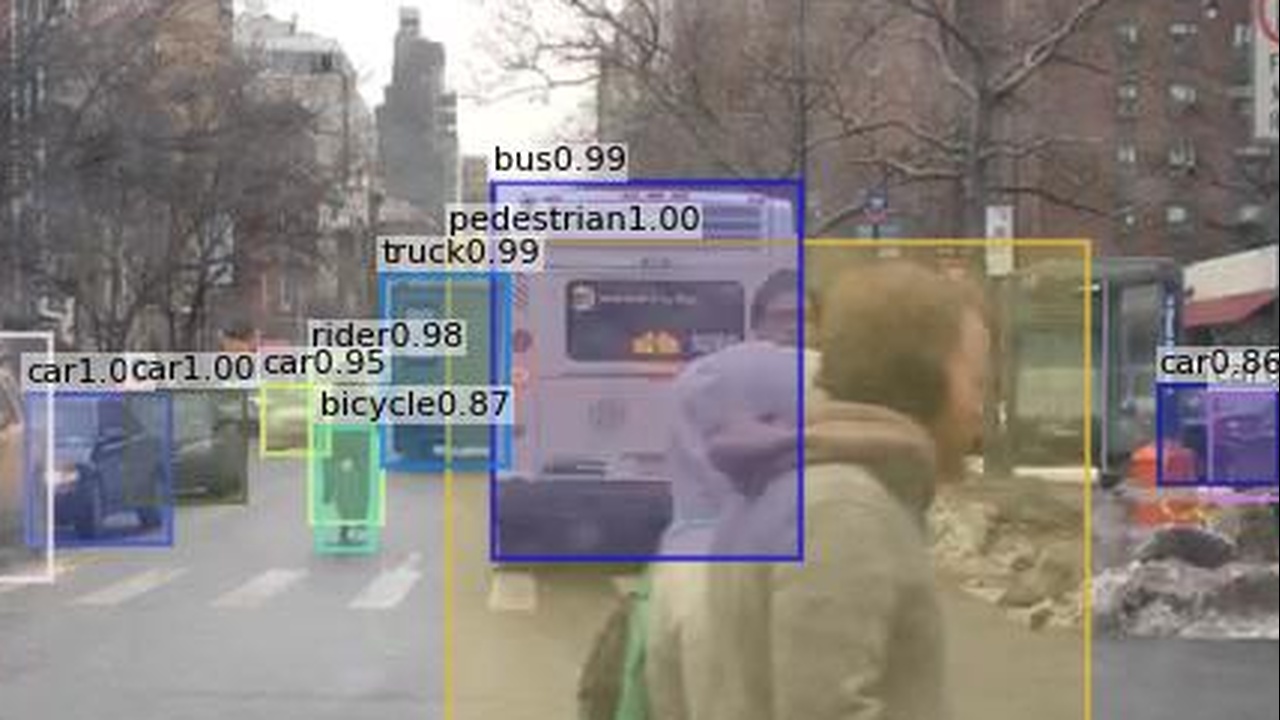}
\\
\includegraphics[width=0.25\linewidth]{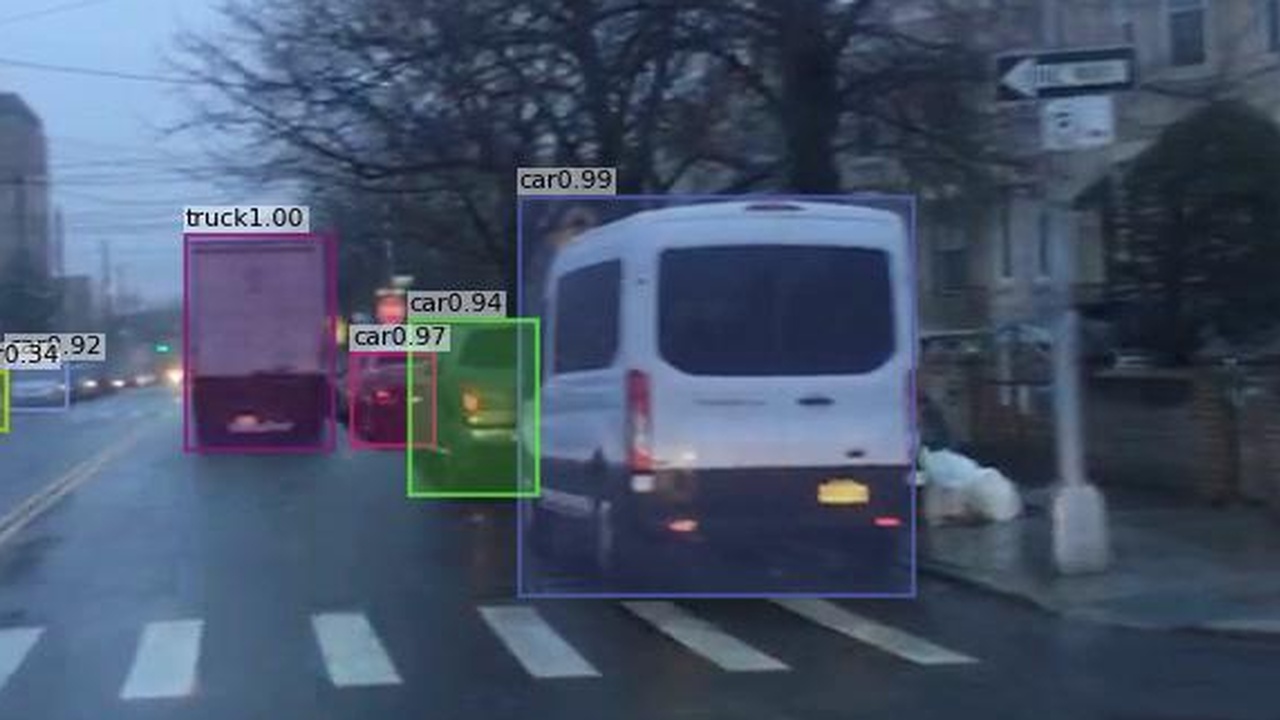} &
\includegraphics[width=0.25\linewidth]{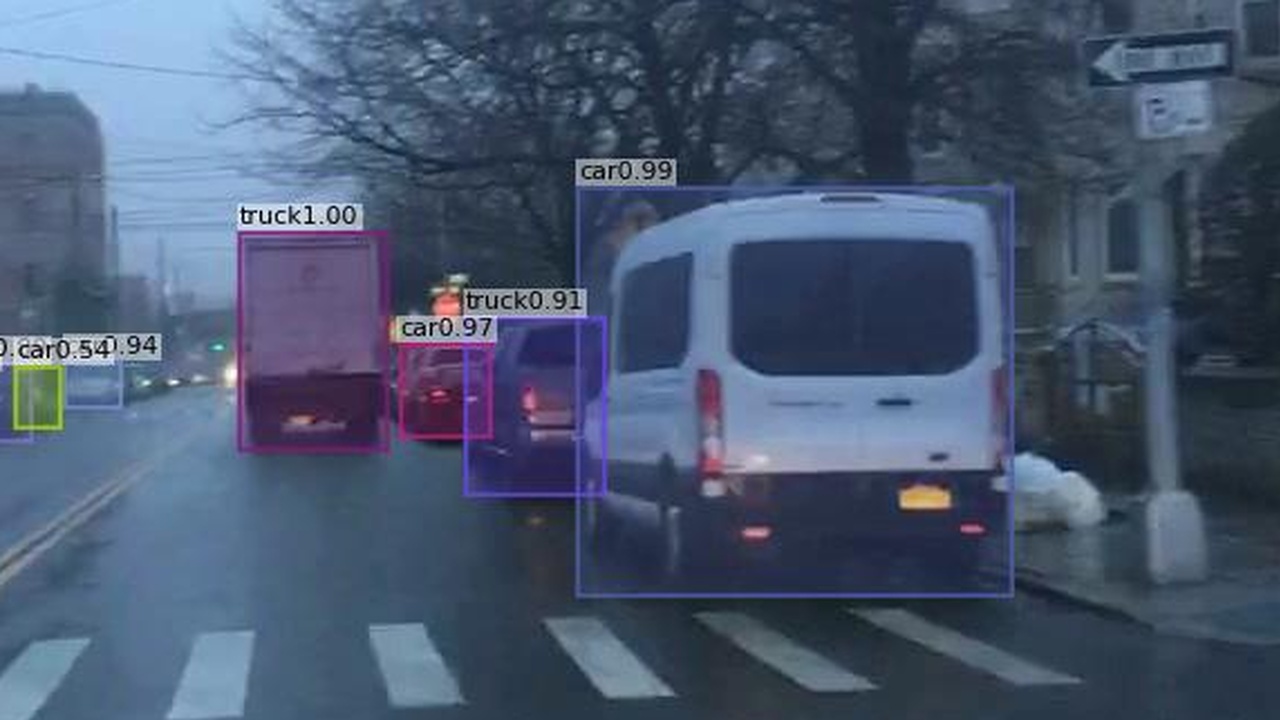} &
\includegraphics[width=0.25\linewidth]{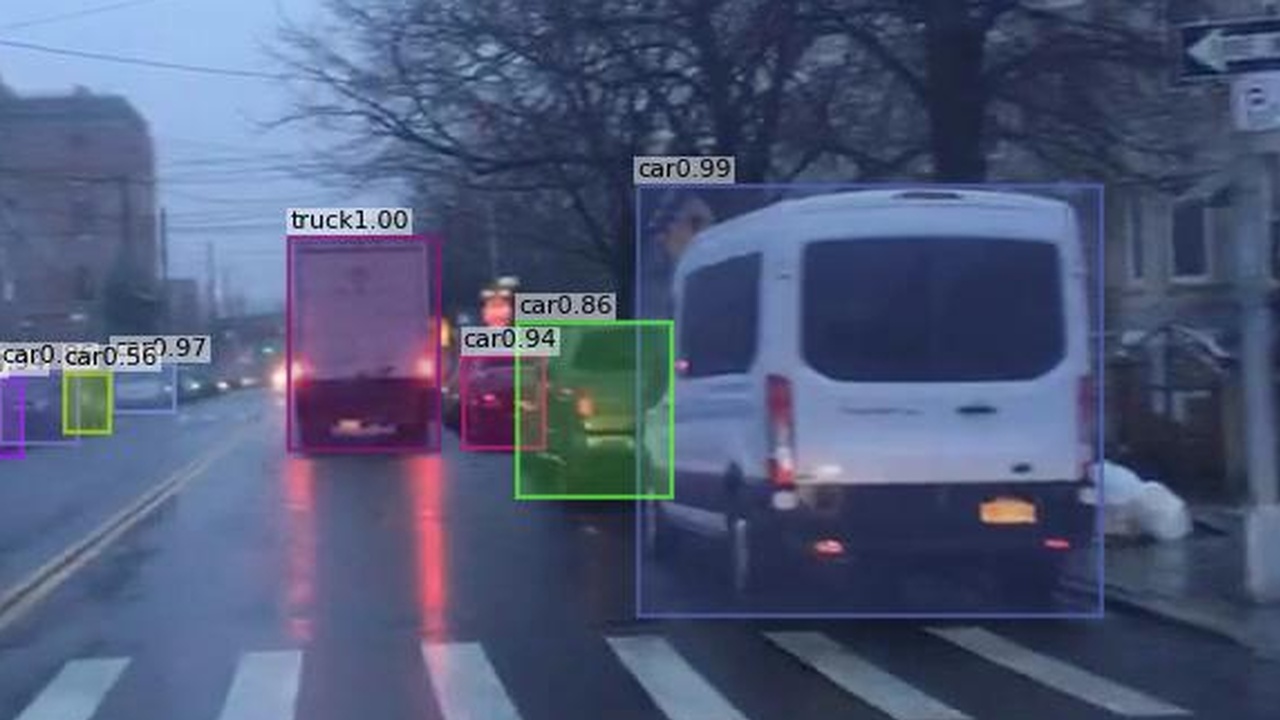} &
\includegraphics[width=0.25\linewidth]{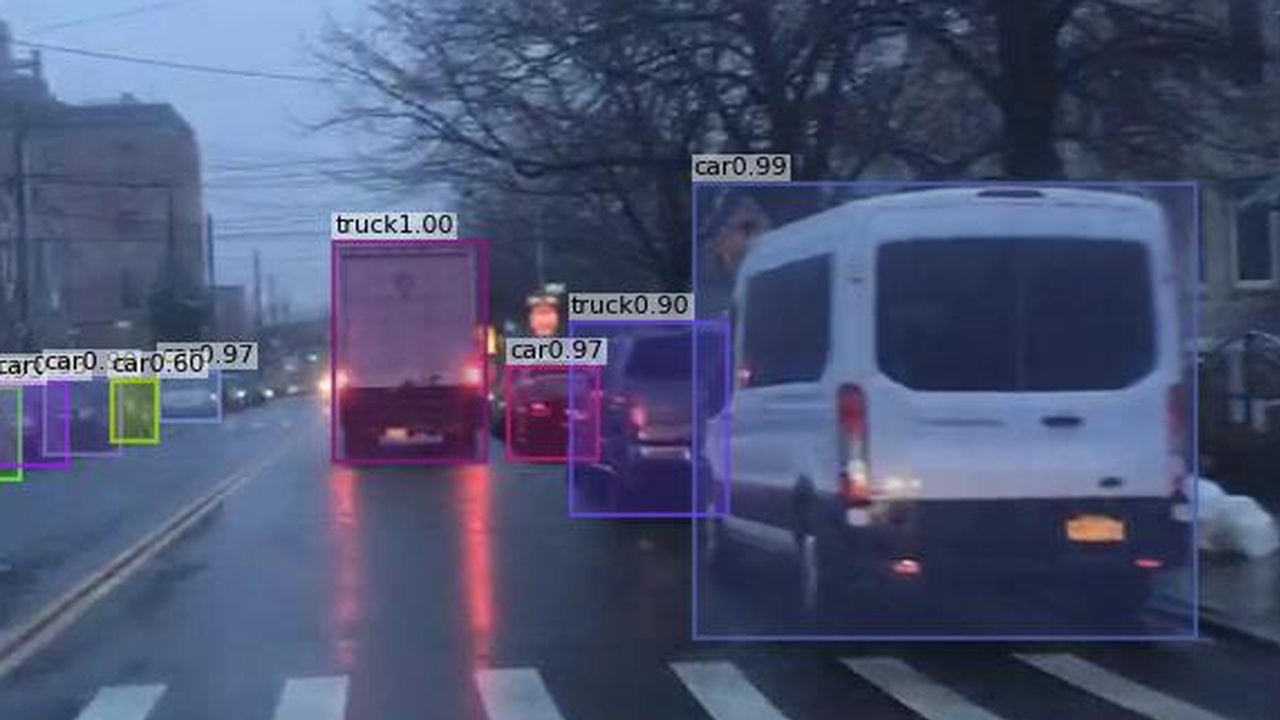} 
\\
\end{tabular}}
\caption{\textbf{Illustration of failure cases.} We illustrate the two most common failure cases of our method (best viewed digitally). In the top, we can see that the bus (\textcolor{lime}{light green} and \textcolor{BlueViolet}{violet}) switches identity due to extreme occlusion by pedestrians. In the bottom, we observe that the pickup truck (\textcolor{green}{green} and \textcolor{Plum}{purple}) switches identity when the class prediction changes between `truck' and `car'. Note that we still re-identify the pickup truck once the class predictions match.}
\label{fig:limitations}
\end{figure*}

%% file: tables/bdd100k_mots.tex
\begin{table}[t]
    \caption{\textbf{Extension of our method to segmentation tracking.} We show segmentation tracking results of our method on the BDD100K segmentation tracking validation set. I: ImageNet. C: COCO. S: Cityscapes. B: BDD100K.}
    \centering
    \resizebox{\linewidth}{!}{
        \begin{tabular}{lccccc}
            \toprule
            Method           & Pretraining & mMOTSA $\uparrow$ & mIDF1 $\uparrow$ & ID sw. $\downarrow$ \\
            \midrule
            SORT~\cite{sort} & I, C, S    & 10.3              & 21.8             & 15951                \\
            MaskTrackRCNN~\cite{trackrcnn} & I, C, S    & 12.3 & 26.2             & 9116                \\
            STEm-Seg~\cite{athar2020stem} & I, C, S    & 12.2 & 25.4             & 8732                \\
            \midrule
            QDTrack (ours)    & I, B       & 25.6 & 45.2             & 980                 \\
            PCAN~\cite{ke2021prototypical}    & I, B       & 27.4 & 45.1             & 876                 \\
            \bottomrule
        \end{tabular}
    }
    \label{tab:bddtrack}
\end{table}

%% file: sections/conclusions.tex

\section{Conclusion}
We present QDTrack, a tracking method based on quasi-dense instance similarity learning.
The key idea behind our method is to utilize all object regions in an image for similarity learning, in contrast to previous methods that only use sparse ground-truth regions as similarity supervision.
We observe that the feature embedding space we learn from quasi-dense matches is much better suited to discriminate instances, allowing for a simple tracking framework that associates objects via nearest neighbor search in the embedding space without bells and whistles. Our method can be easily coupled with most existing object detectors and feature extractors for end-to-end training, and learns effective instance similarity even without video input or tracking annotations.

%% file: sections/appendix.tex
\appendices

\section{Additional details}

We detail the training settings and hyper-parameters used for each benchmark investigated in the main paper.
We also provide details regarding the augmentations used during training.

\paragraph{Training setting}
On MOT17 and MOT20, we train QDTrack with YOLOX~\cite{ge2021yolox} on the union of CrowdHuman~\cite{shao2018crowdhuman} and the respective MOT benchmark for 80 epochs, following recent practice~\cite{zhang2021bytetrack, cao2022observation, du2022strongsort}.
We use an image scale of $1440\times800$ for MOT17 and $1600\times896$ for MOT20.
We use a batch size of 32 with a learning rate of 0.0005, and
we use a cosine annealing learning rate schedule, ending at a learning rate 0.05 times the original.
We use an exponential learning rate warm-up for one epoch.
For augmentations, we turn off MixUp and Mosaic for the last ten epochs.
We also use an exponential moving average (EMA), as done in~\cite{ge2021yolox}.
On DanceTrack, we use the same training setup as in MOT17, except for training for 12 epochs.
On BDD100K, we use mostly the same training setup as in MOT17, except for training for 25 epochs, using a batch size of 48, learning rate of 0.00075, and not turning off augmentations.

\paragraph{Hyper-parameters}
The detailed hyper-parameters are shown in Table~\ref{tab:hyperparams}.
As our object association only relies on appearance, it is robust to different motion patterns in different datasets.
The experiments share similar tracking parameters except for TAO, since TAO uses 3D mAP instead of CLEAR MOT and HOTA metrics for evaluation.
On TAO, $\beta_{\texttt{new}}$ and $\beta_{\texttt{obj}}$ are set to 0.0001 to obtain a high recall.
Considering the numerous tracks that results using these thresholds, we do not maintain backdrops.

\paragraph{Augmentation parameters}
For Mosaic, we sample the center of the mosaic image in the range (0.5, 1.5).
For the main results on the benchmarks, we also use random affine transformation with a rotation degree in the range (-10.0, 10.0), translation factor in the range (-0.1, 0.1), scale factor in the range (0.5, 1.5), and shear degree in the range (-2.0, 2.0).
For MixUp, we jitter the additional image by a factor in the range (0.5, 1.5) and flip it with a probability of 0.5.
We also flip the resulting combined image with a probability of 0.5.
We additionally apply random resizing with a scale range (0.5, 1.5) while maintaining the aspect ratio and random cropping.
Finally, we apply color jitter with a brightness factor in the range (0.875, 1.125), contrast factor in the range (0.5, 1.5), saturation factor in the range (0.5, 1.5), and hue shift in the range ($-0.2\pi, 0.2\pi$).

\section{Additional details for MOT Challenge}
For challenging scenes with heavy occlusions (\ie, MOT17 and MOT20), object association with only appearance cues can be very challenging, as there are large overlaps between objects.
We utilize several additional techniques to mitigate these issues.
First, object appearance cues immediately after re-appearance can be unreliable for association, leading to a higher number of ID switches and lower \texttt{AssRe} score.
We address this by introducing a near-online merging strategy.
For each object that did not match to any previous tracks and initialized a new track, in each of the $t$ subsequent frames we merge its current track with a vanished track if their matching score is higher than a threshold $\beta_{\texttt{merge}}$.
We use distance thresholding with distance $d_{\texttt{merge}}$ to ignore objects that are too far away.
This enables us to utilize more reliable appearance features moments after re-appearance for matching.
We use $t=10$, $\beta_{\texttt{merge}}=0.5$, $d_{\texttt{merge}}=50$.

\setlength{\tabcolsep}{2pt}
\begin{table}[h]
    \caption{\textbf{Hyper-parameters used in each benchmark.} We include both tracking and detection parameters.}
    \scriptsize
    \centering
    \resizebox{\linewidth}{!}{
        \begin{tabular}{ccccccccccccc}
            \toprule
            Parameter & MOT17 & MOT20 & DanceTrack & BDD100K & Waymo & TAO \\
            \midrule
            $\beta_{\texttt{new}}$ & 0.75 & 0.75 & 0.8 & 0.5 & 0.8 & 0.0001 \\
            $\beta_{\texttt{obj}}$ & 0.3 & 0.3 & 0.6 & 0.35 & 0.5 & 0.0001 \\
            $\beta_{\texttt{match}}$ & 0.5 & 0.5 & 0.5 & 0.5 & 0.5 & 0.5 \\
            $K$ & 30 & 30 & 20 & 10 & 10 & 10 \\
            $L$ & 1 & 1 & 1 & 1 & 1 & - \\
            $m$ & 0.5 & 0.5 & 0.8 & 0.8 & 0.8 & 0.8 \\
            Det. confidence & 0.1 & 0.001 & 0.1 & 0.1 & 0.05 & 0.0001 \\
            Det. NMS threshold & 0.7 & 0.7 & 0.7 & 0.65 & 0.7 & 0.5 \\
            \bottomrule
        \end{tabular}
    }
    \label{tab:hyperparams}
\end{table}
\setlength{\tabcolsep}{6pt}

\begin{table}[h]
    \caption{\textbf{Ablation study of association strategies used for MOT Challenge.} We evaluate on the MOT17 validation set.}
    \scriptsize
    \centering
    \resizebox{\linewidth}{!}{
        \begin{tabular}{cccccc}
            \toprule
            Distance & Tracklet & Linear & \multirow{2}{*}{MOTA $\uparrow$} & \multirow{2}{*}{IDF1 $\uparrow$} & \multirow{2}{*}{HOTA $\uparrow$}  \\
            Threshold & Merging & Interpolation & & & \\
            \midrule
            - & - & -                               & 76.1 & 73.6 & 63.5 \\
            \checkmark & - & -                     & 76.2 & 74.5 & 64.0 \\
            \checkmark & \checkmark & -           & 76.0 & 76.0 & 64.4 \\
            \checkmark & \checkmark & \checkmark & \textbf{76.8} & \textbf{76.2} & \textbf{64.8} \\
            \bottomrule
        \end{tabular}
    }
    \label{tab:sup-assoc}
\end{table}

\begin{table}[h]
    \caption{\textbf{Ablation study of momentum of the embeddings.} We evaluate on the BDD100K tracking validation set.}
    \scriptsize
    \centering
    \resizebox{\linewidth}{!}{
        \begin{tabular}{ccccccccccc}
            \toprule
            Momentum & mMOTA $\uparrow$ & mIDF1 $\uparrow$ & MOTA $\uparrow$ & IDF1 $\uparrow$ \\
            \midrule
            0.6      & 37.0             & 50.9             & 63.3            & 71.4            \\
            0.7      & 37.0             & 50.9             & 63.3            & 71.3            \\
            0.8      & 37.0             & 50.7             & 63.3            & 71.1            \\
            0.9      & 37.0             & 50.6             & 63.3            & 70.8            \\
            1.0      & 37.0             & 50.5             & 63.3            & 70.5            \\
            \bottomrule
        \end{tabular}
    }
    \label{tab:momentum}
\end{table}

Additionally, due to the high frame rate and low object motion in the MOT benchmarks, we use distance thresholding to reduce ambiguities during association by ignore matching candidates that are greater than a distance $d$ away.
We use $d=50$.
Following~\cite{zhang2021bytetrack}, we also perform linear interpolation to recover bounding boxes of fully-occluded objects.

We provide an ablation study of the aforementioned techniques on the MOT benchmarks, MOT17 and MOT20.
The results on the MOT17 validation set are shown in Table~\ref{tab:sup-assoc}.
Using distance thresholding can improve IDF1 from 73.6 to 74.5 (+0.9).
Performing tracklet merging can improve IDF1 from 74.5 to 76.0 (+1.5).
Linear interpolation can further improve all metrics.



\begin{table*}[t]
    \caption{\textbf{Detection oracle analysis.} The numbers in the round brackets mean the gaps between oracle results and our results.}
    \centering
    \small
    \setlength\tabcolsep{3.8mm}
    \begin{tabular}{lcccccccc}
        \toprule
        Category  & MOTA $\uparrow$ & IDF1 $\uparrow$ & MOTP $\uparrow$ & FN $\downarrow$ & FP $\downarrow$ & ID Sw. $\downarrow$ & MT $\uparrow$ & ML $\downarrow$ \\
        \midrule
        Pedestrian & 94.3            & 79.5 (+19.3)    & 99.8            & 1               & 1               & 3226                & 3506          & 0               \\
        Rider      & 95.8            & 88.5 (+40.4)    & 99.9            & 0               & 0               & 107                 & 134           & 0               \\
        Car        & 97.7            & 86.1 (+11.1)    & 99.9            & 0               & 0               & 7716                & 13189         & 0               \\
        Bus        & 99.2            & 93.0 (+31.2)    & 100.0           & 0               & 0               & 72                  & 196           & 0               \\
        Truck      & 98.8            & 90.3 (+33.8)    & 100.0           & 0               & 0               & 340                 & 726           & 0               \\
        Bicycle    & 88.2            & 79.5 (+31.8)    & 98.7            & 8               & 8               & 470                 & 243           & 0               \\
        Motorcycle & 97.0            & 94.5 (+37.8)    & 99.8            & 0               & 0               & 27                  & 44            & 0               \\
        Train      & 99.4            & 98.7 (+98.7)    & 100.0           & 0               & 0               & 2                   & 6             & 0               \\
        \midrule
        All        & 96.3            & 88.8 (+38.0)    & 99.8            & 9               & 9               & 11960               & 18044         & 0               \\
        \bottomrule
    \end{tabular}
    \label{tab:oracle}
\end{table*}

\begin{table*}[t]
    \caption{\textbf{Tracking oracle analysis.} The numbers in the round brackets mean the gaps between oracle results and our results.}
    \centering
    \small
    \setlength\tabcolsep{3.6mm}
    \begin{tabular}{lcccccccc}
        \toprule
        Category  & MOTA $\uparrow$ & IDF1 $\uparrow$ & MOTP $\uparrow$ & FN $\downarrow$ & FP $\downarrow$ & ID Sw. $\downarrow$ & MT $\uparrow$ & ML $\downarrow$ \\
        \midrule
        Pedestrian & 54.7            & 71.2 (+11.0)    & 77.6            & 14990           & 10095           & 755                 & 1835          & 367             \\
        Rider      & 31.4            & 52.6 (+4.5)     & 76.6            & 1390            & 242             & 115                 & 16            & 56              \\
        Car        & 74.3            & 82.9 (+7.9)     & 84.1            & 54585           & 31014           & 2309                & 8759          & 1141            \\
        Bus        & 38.2            & 65.8 (+4.0)     & 86.1            & 3532            & 2031            & 57                  & 61            & 41              \\
        Truck      & 37.0            & 60.9 (+4.4)     & 84.7            & 12719           & 4259            & 247                 & 149           & 239             \\
        Bicycle    & 30.6            & 55.6 (+7,9)     & 75.4            & 2031            & 714             & 125                 & 60            & 58              \\
        Motorcycle & 14.6            & 51.7 (-5.0)     & 76.4            & 443             & 292             & 35                  & 10            & 18              \\
        Train      & -0.6            & 0.0 (+0.0)      & 0.0             & 308             & 2               & 0                   & 0             & 6               \\
        \midrule
        All        & 35.0            & 55.1 (+4.3)     & 70.1            & 89998           & 48649           & 3643                & 10890         & 1926            \\
        \bottomrule
    \end{tabular}
    \label{tab:track_oracle}
\end{table*}

\section{Additional ablation studies}

\paragraph{Momentum of the embeddings}
Assume there is an existing track and its embedding is $E_0$.
This track is associated to an object on the current frame and its embedding is $E_1$.
The new embedding of this track will be $m * E_1 + (1-m) * E_0$, where $m$ is the momentum. The momentum does not improve the results too much but it considers the history of embeddings.
We show the ablation studies of different values of momentum in Table \ref{tab:momentum}.
The models for this table are re-trained so the results are slightly different from the results in the main paper.


\paragraph{Sensitivity of $\gamma_1$ and $\gamma_2$ in Eq.~\ref{eqa:loss}}
We found $\gamma_2$ does not change the final results while $\gamma_1$ does. If $\gamma_1$ is higher than 0.5, the performance will drop, but does not matter if it is lower than 0.5.

\section{Oracle analysis}

We investigate the performances of two types of oracles on the BDD100K tracking validation set: detection oracle and tracking oracle.
For the detection oracle, we directly extract feature embeddings of the ground truth objects in each frame and associate them using our method.
For the tracking oracle, we use ground truth tracking labels to associate the detected objects.

\paragraph{Detection oracle}
The results are shown in Table~\ref{tab:oracle}.
We can observe that all MOTAs are higher than 94\%, and some of them are even close to 100\%.
This is because we use the ground truth boxes directly so that the number of false negatives and false positives are close to 0.

The metric IDF1 and ID Switches can measure the performance of identity consistency.
The average IDF1 over the 8 classes is 88.8\%, which is 38 points higher than our result.
The gaps on classes ``car'' and ``pedestrain'' are only 11.1 points and 19.3 points between oracle results and our results respectively, while gaps on other classes are exceeding 30 points.
These results show that if highly accurate detection results are provided, our method can obtain robust feature embeddings and associate objects effectively.
However, the huge performance gaps also indicate the demand of promoting detection algorithms in the video domain.
We also notice that the total number of ID switches in the oracle experiment is higher than ours.
This is due to the high object recalls in the oracle experiments, as more detected instances may introduce more ID switches accordingly.

\paragraph{Tracking oracle}
The results are shown in Table~\ref{tab:track_oracle}.
We can observe that when associating object directly with tracking labels, the mIDF1 is only boosted by 4.3 points.
This promising oracle analysis shows the effectiveness of our method and indicates that our method is bounded more by detection performance than tracking performance.

\section{Failure case analysis}
Our method can distinguish different instances even they are similar in appearance.
However, there are still some failure cases.
We show them below with figures, in which we use yellow color to represent false negatives, red color to represent false positives, and cyan color to represent ID switches.
The float number at the corner of each box indicates the detection score, while the integer indicates the object identity number.
We use green dashed box to highlight the objects we want to emphasize.

\begin{figure*}[h!]
    \centering
    \begin{subfigure}[b]{0.49\textwidth}
        \centering
        \includegraphics[width=\textwidth]{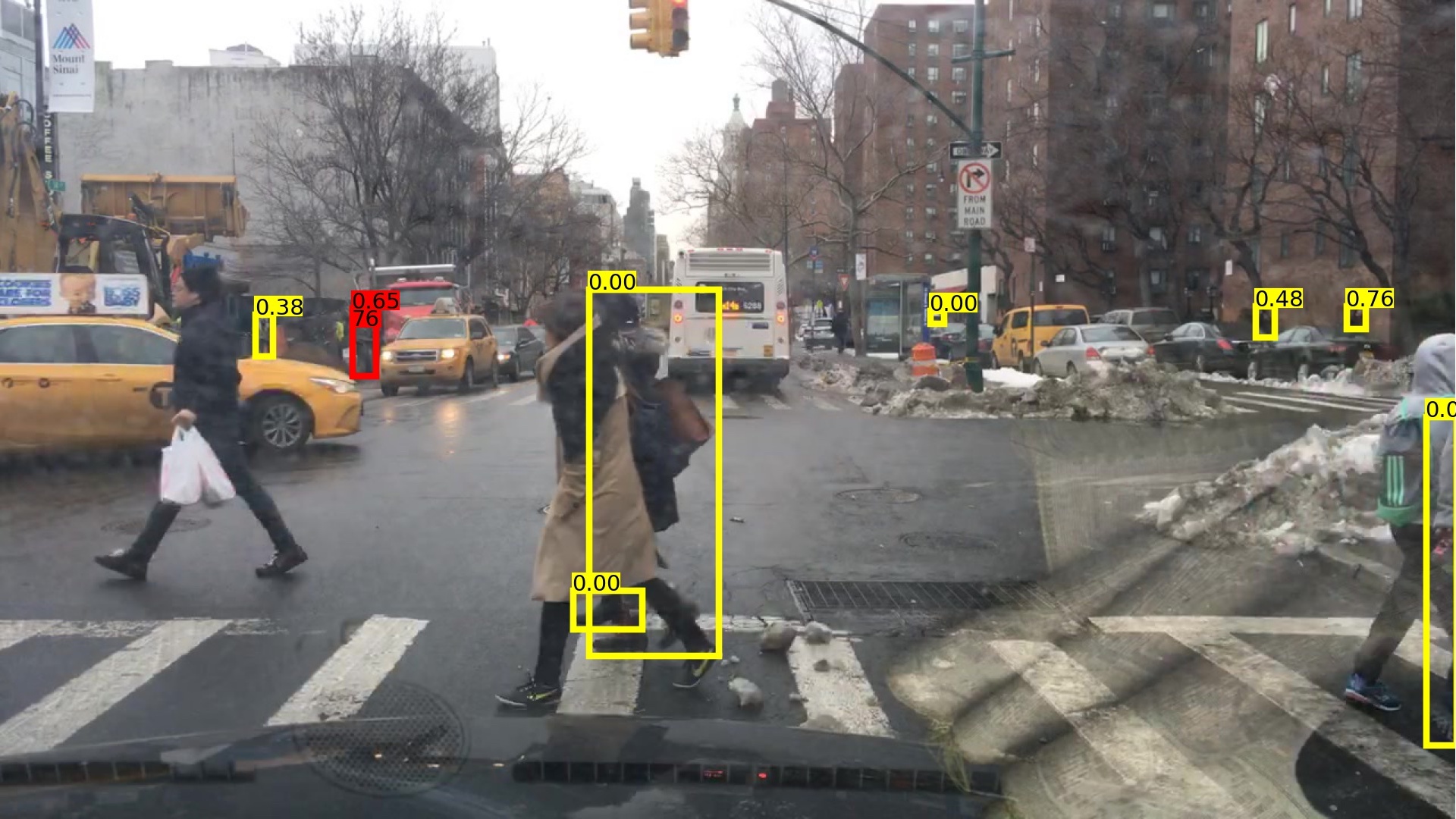}
    \end{subfigure}
    \begin{subfigure}[b]{0.49\textwidth}
        \centering
        \includegraphics[width=\textwidth]{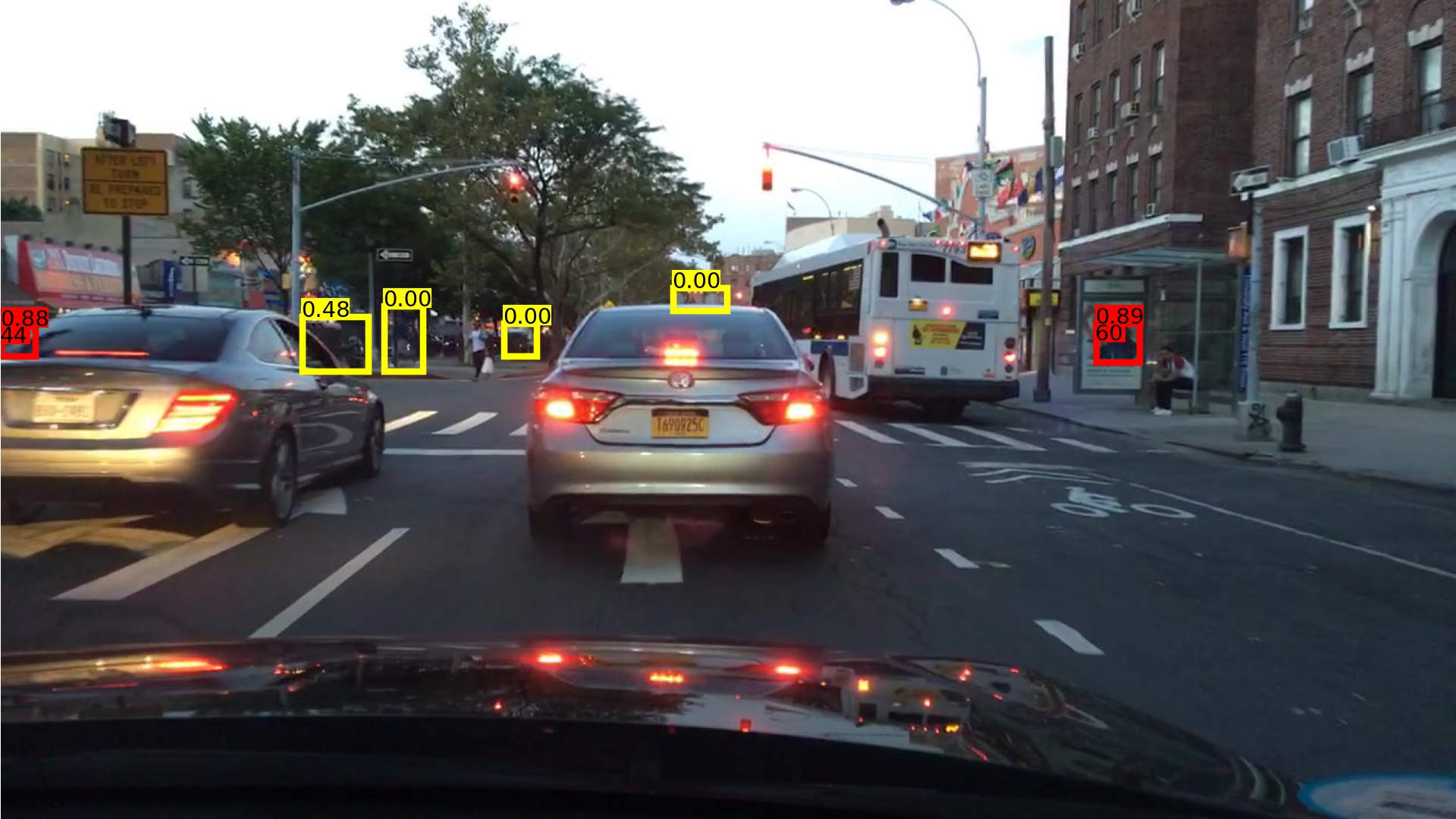}
    \end{subfigure}
    \caption{\textbf{Failure cases caused by inaccurate classification confidences.} The objects enclosed by yellow rectangles are false negatives, and the objects enclosed by red rectangles are false positives.}
    \label{fig:fp+fn}
\end{figure*}

\paragraph{Object classification}
Inaccurate classification confidence is the main distraction for the association procedure because false negatives and false positives destroy the one-to-one matching constraint.
As shown in Figure~\ref{fig:fp+fn}, the false negatives are mainly small objects or occluded objects under crowd scenes.
The false positives are objects that have similar appearances to annotated objects, such as persons in the mirror or advertising board, etc.

\begin{figure*}
    \centering
    \begin{subfigure}[b]{0.49\textwidth}
        \centering
        \includegraphics[width=\textwidth]{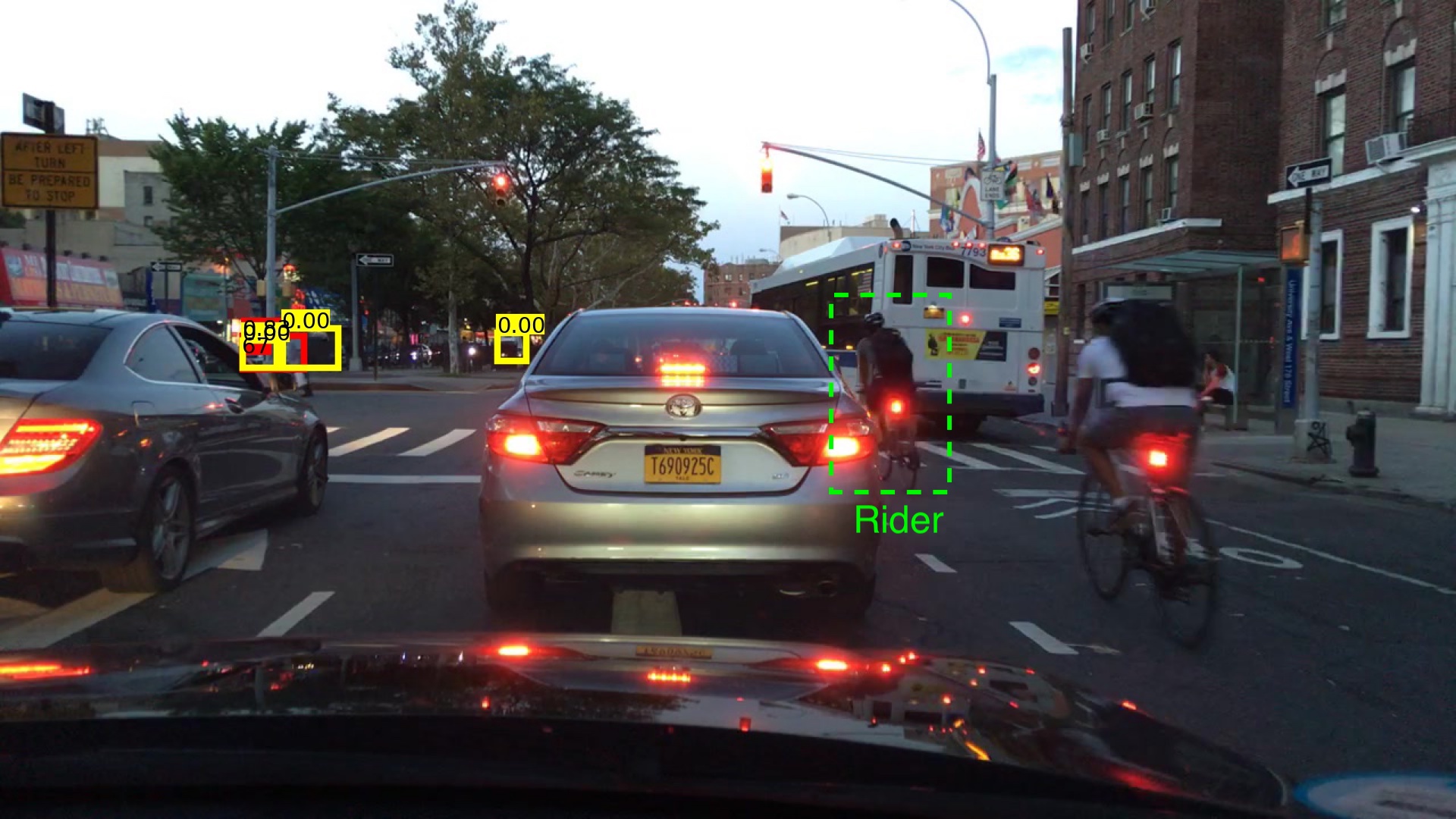}
    \end{subfigure}
    \begin{subfigure}[b]{0.49\textwidth}
        \centering
        \includegraphics[width=\textwidth]{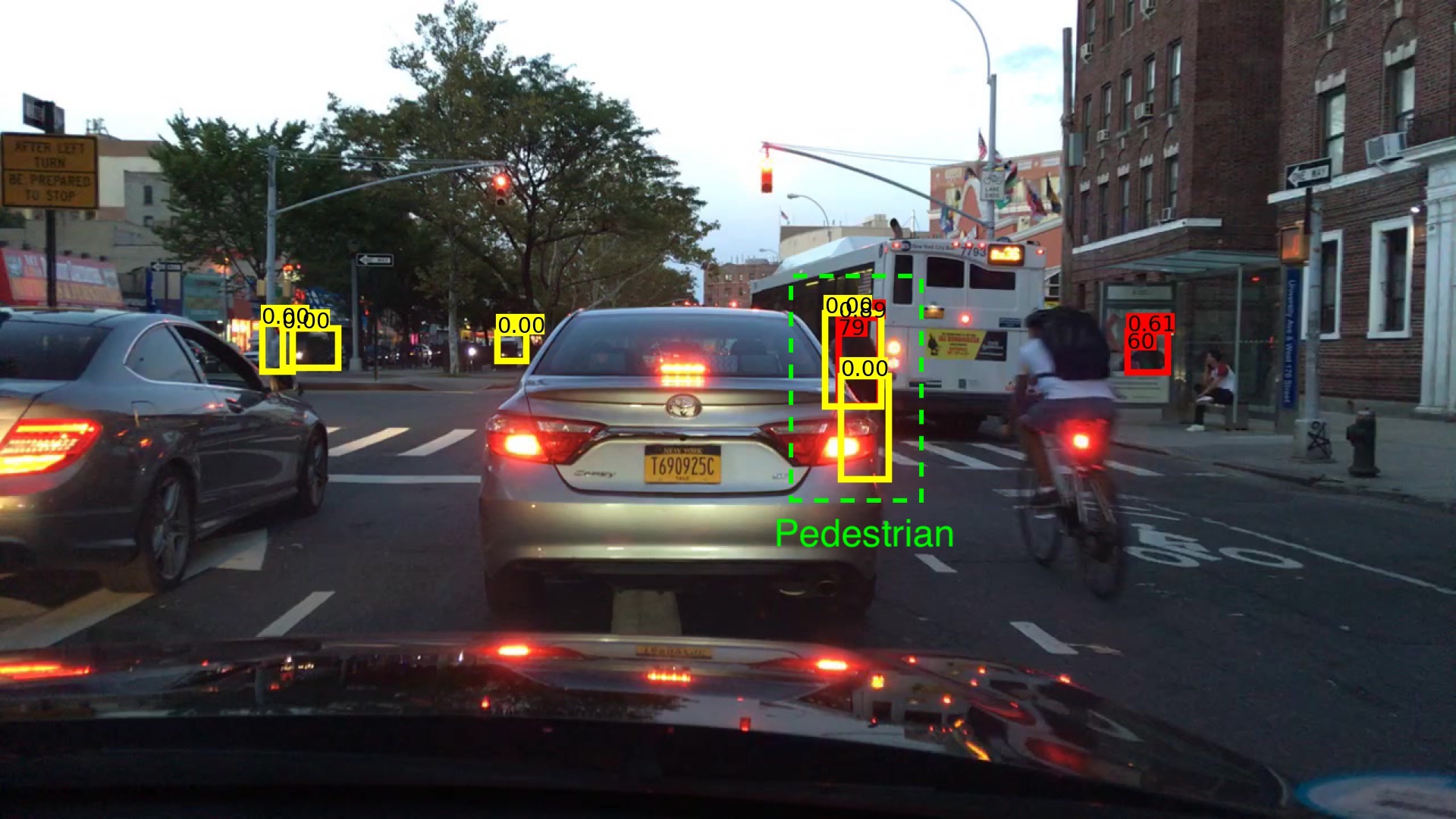}
    \end{subfigure}
    \caption{\textbf{Failure case caused by inaccurate object category.} The category of the highlighted object changes from ``rider'' to ``pedestrian'' due to the occlusion of the bicycle. They cannot be associated because they do not satisfy the category consistency.}
    \label{fig:det_cat}
\end{figure*}

Inaccurate object category is a less frequent distraction caused by classification.
The class of the instance may switch between different categories, which mostly belong to the same super-category.
Figure~\ref{fig:det_cat} shows an example.
The category of the highlighted object changes from ``rider'' to ``pedestrian'' when the bicycle is occluded.
Our method fails in this case because we require the associated objects have the same category.

These failure cases caused by object classification suggest that improvements could be achieved via leveraging video object detection algorithms, \ie exploiting temporal information to improve the detector, thus obtaining better tracking performance.

\begin{figure*}[h!]
    \centering
    \begin{subfigure}[b]{0.49\textwidth}
        \centering
        \includegraphics[width=\textwidth]{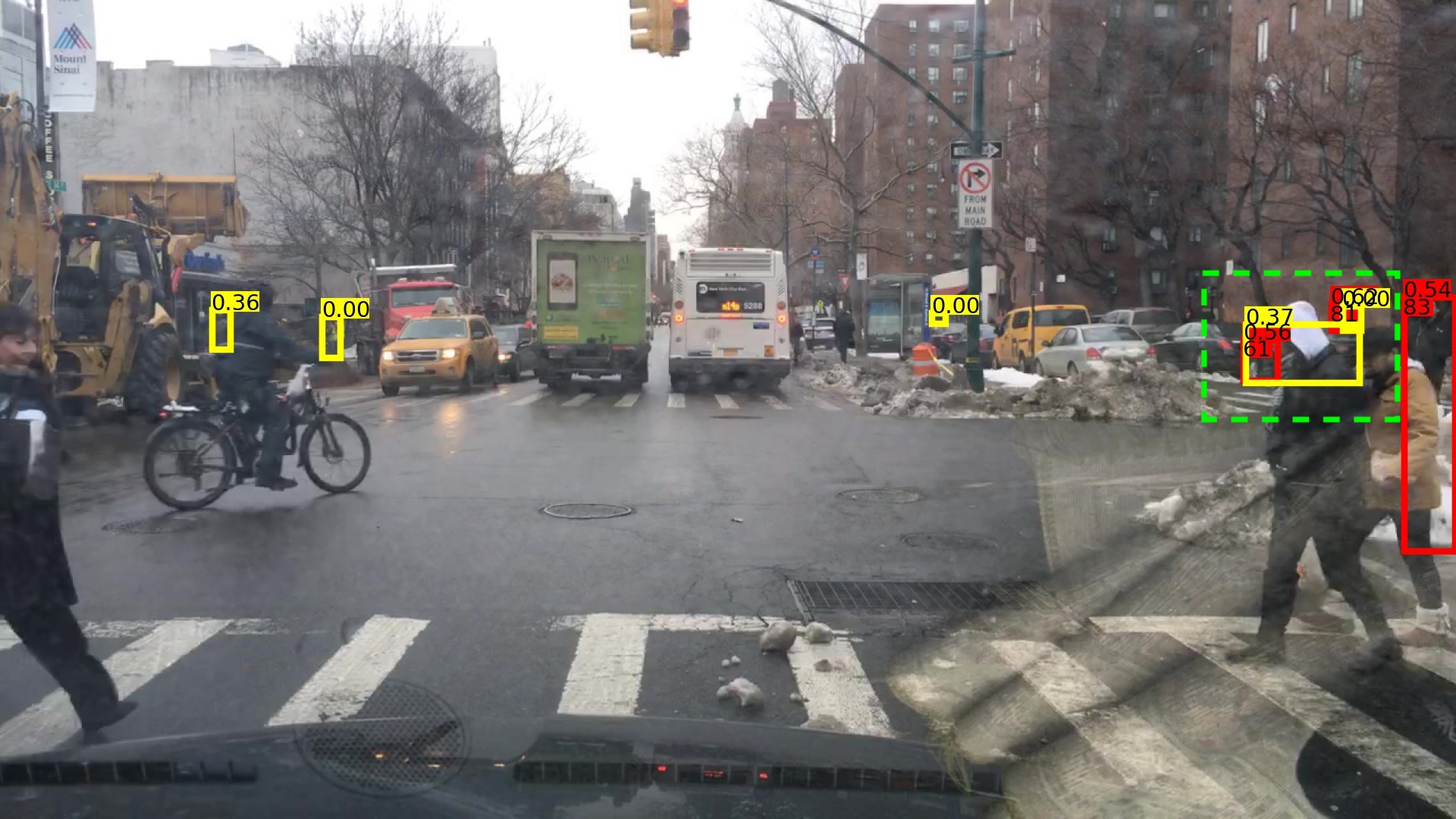}
    \end{subfigure}
    \begin{subfigure}[b]{0.49\textwidth}
        \centering
        \includegraphics[width=\textwidth]{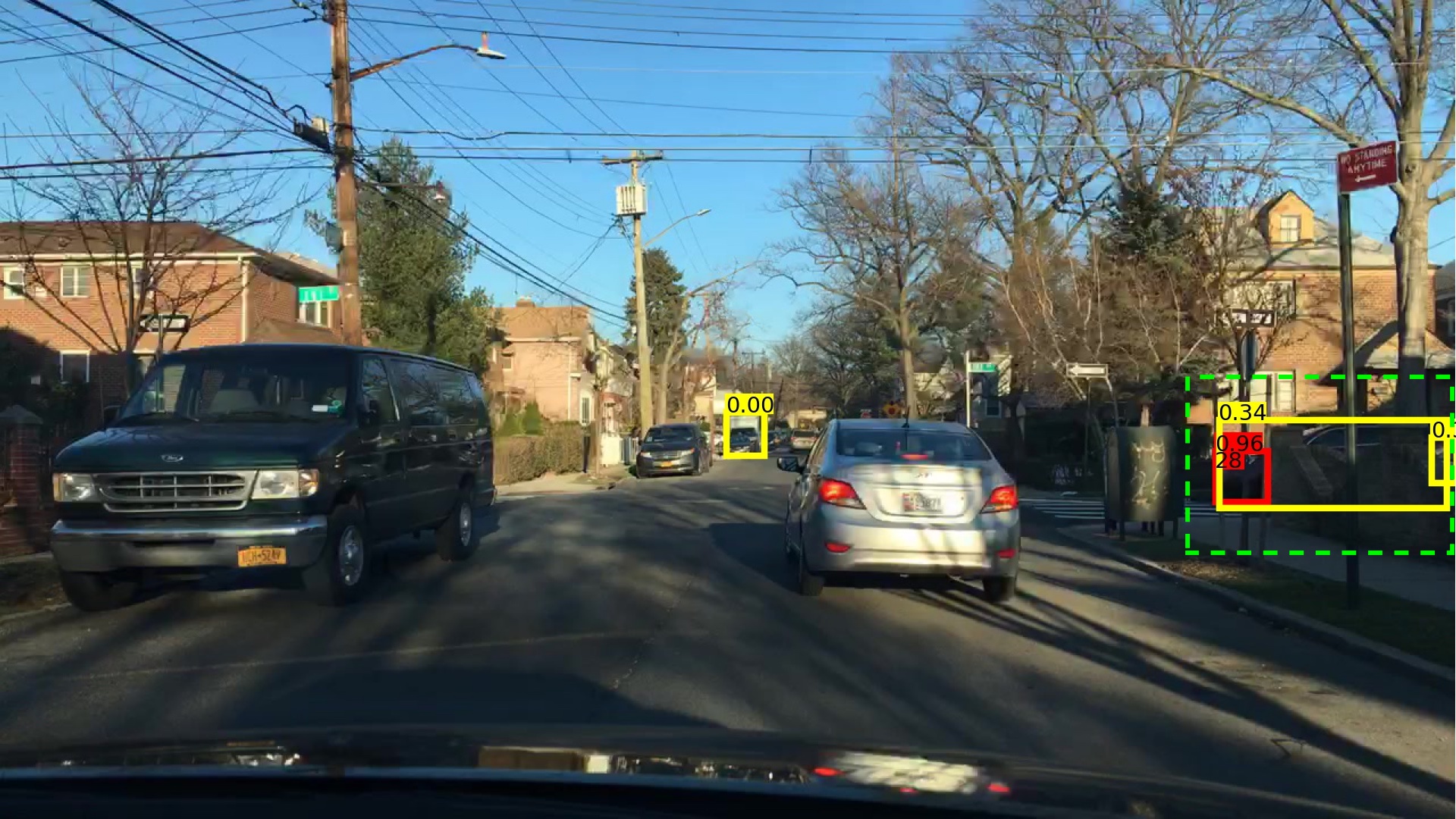}
    \end{subfigure}
    \caption{\textbf{Inaccurate object localization caused by truncation.} The red false positive box only covers part of the object, while the yellow box covers the entire object. They may have similar feature embeddings thus influencing the association procedure. }
    \label{fig:det_loc}
\end{figure*}

\paragraph{Object truncation/occlusion}
Object truncation/occlusion causes inaccurate object localization.
As shown in Figure~\ref{fig:det_loc}, the highlighted objects are truncated by other objects.
The detector detects two objects.
One of them is a false positive box that only covers a part of the object.
The other one is a box with a lower detection score but covers the entire object.
This case may influence the association process if the two boxes have similar feature embeddings.

\begin{figure*}[h!]
    \centering
    \begin{subfigure}[b]{0.49\textwidth}
        \centering
        \includegraphics[width=\textwidth]{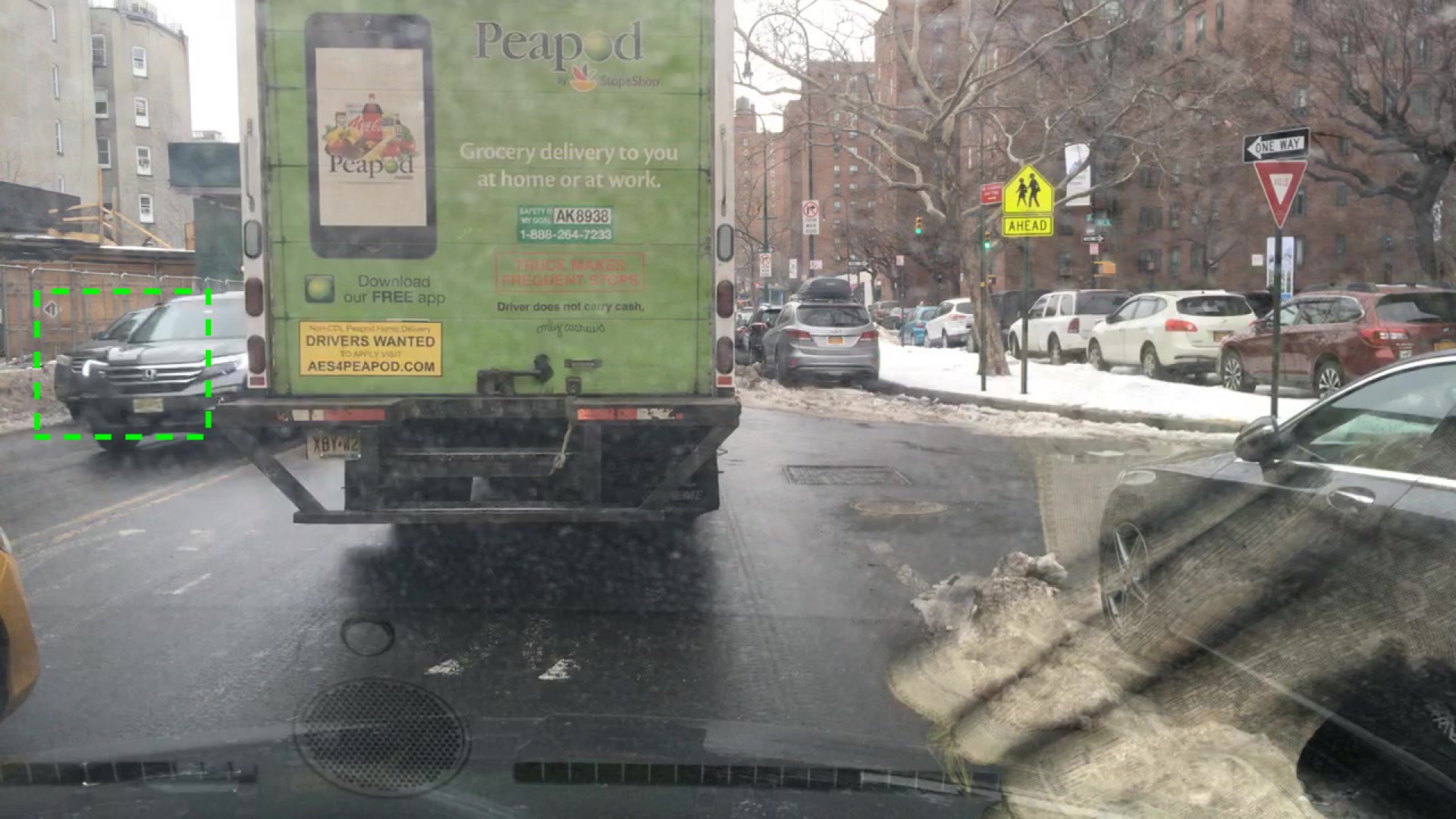}
    \end{subfigure}
    \begin{subfigure}[b]{0.49\textwidth}
        \centering
        \includegraphics[width=\textwidth]{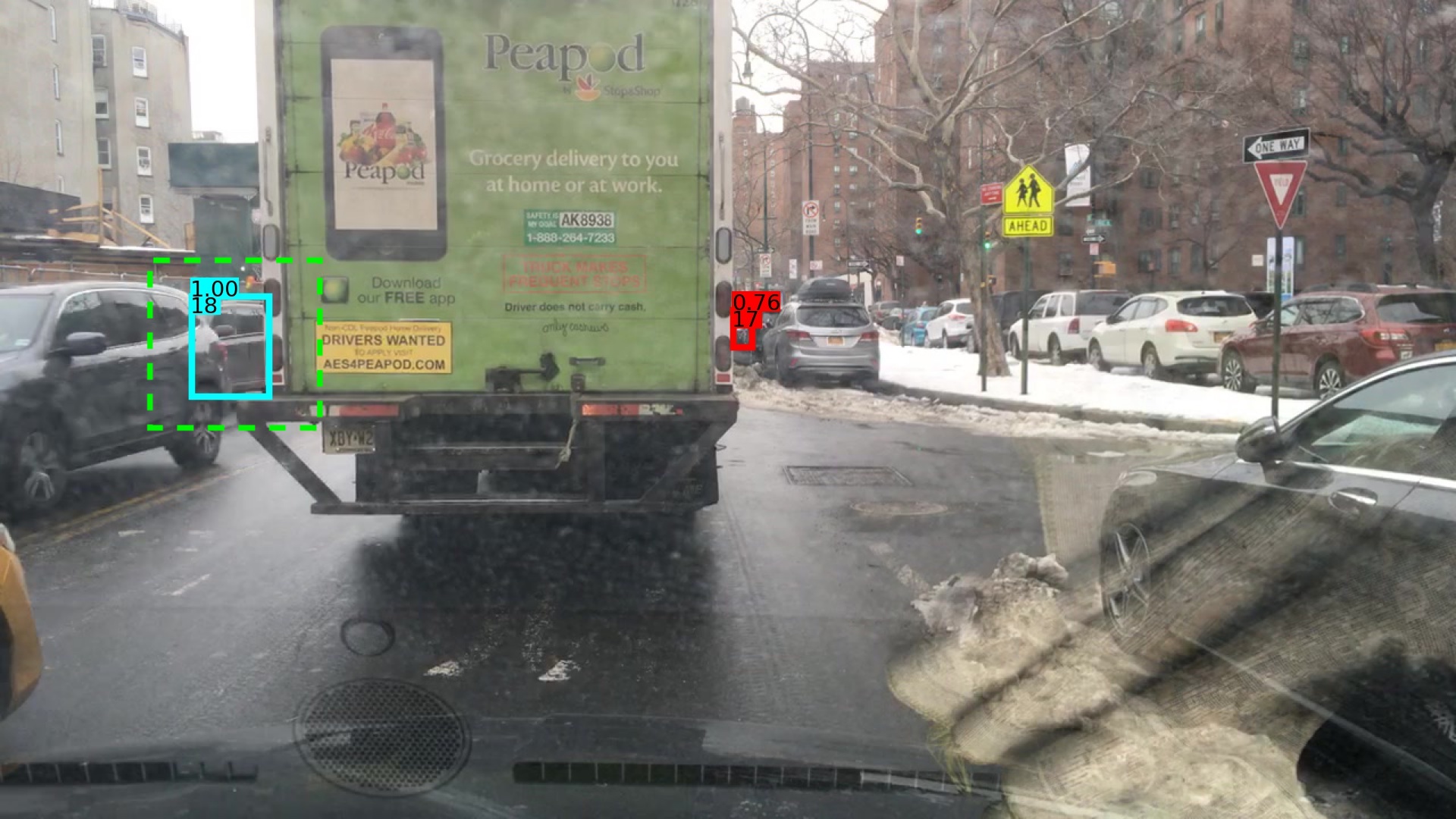}
    \end{subfigure}
    \caption{\textbf{Occlusion in different regions of the same object.} Two detected objects in different frames cover totally different regions of the object thus having low appearance similarity.}
    \label{fig:truncate2}
\end{figure*}

An instance may have totally different appearances before and after occlusion that result in low similarity scores.
As shown in Figure~\ref{fig:truncate2}, only the front of the car appears before occlusion, while only the rear of the car appears after occlusion.
Our method can associate two boxes if they cover the same discriminative regions of an object, not necessarily the exact same region.
However, if two boxes cover totally different regions of the object, they will have a low matching score.

\begin{figure*}
    \centering
    \begin{subfigure}[b]{0.49\textwidth}
        \centering
        \includegraphics[width=\textwidth]{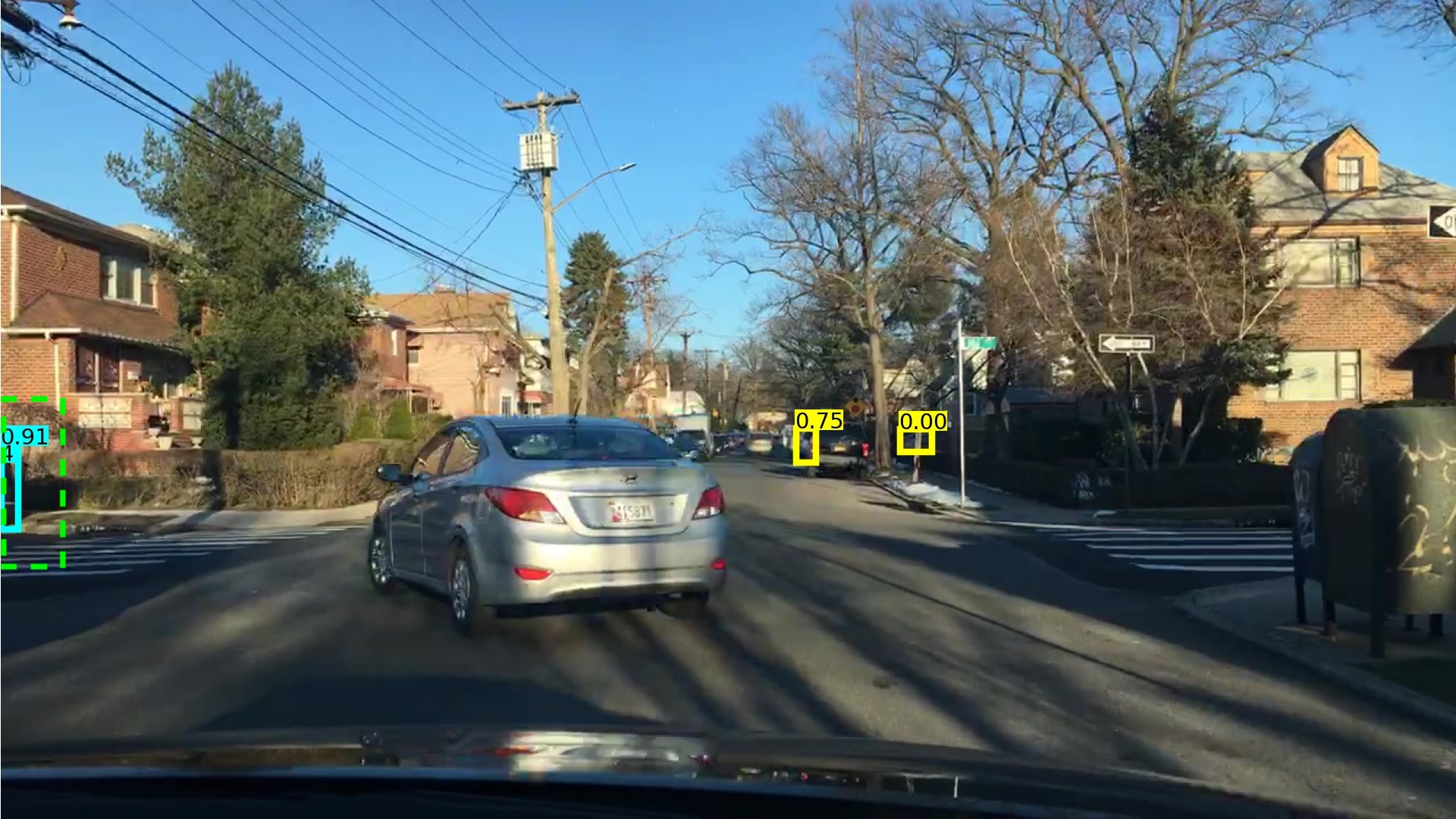}
    \end{subfigure}
    \begin{subfigure}[b]{0.49\textwidth}
        \centering
        \includegraphics[width=\textwidth]{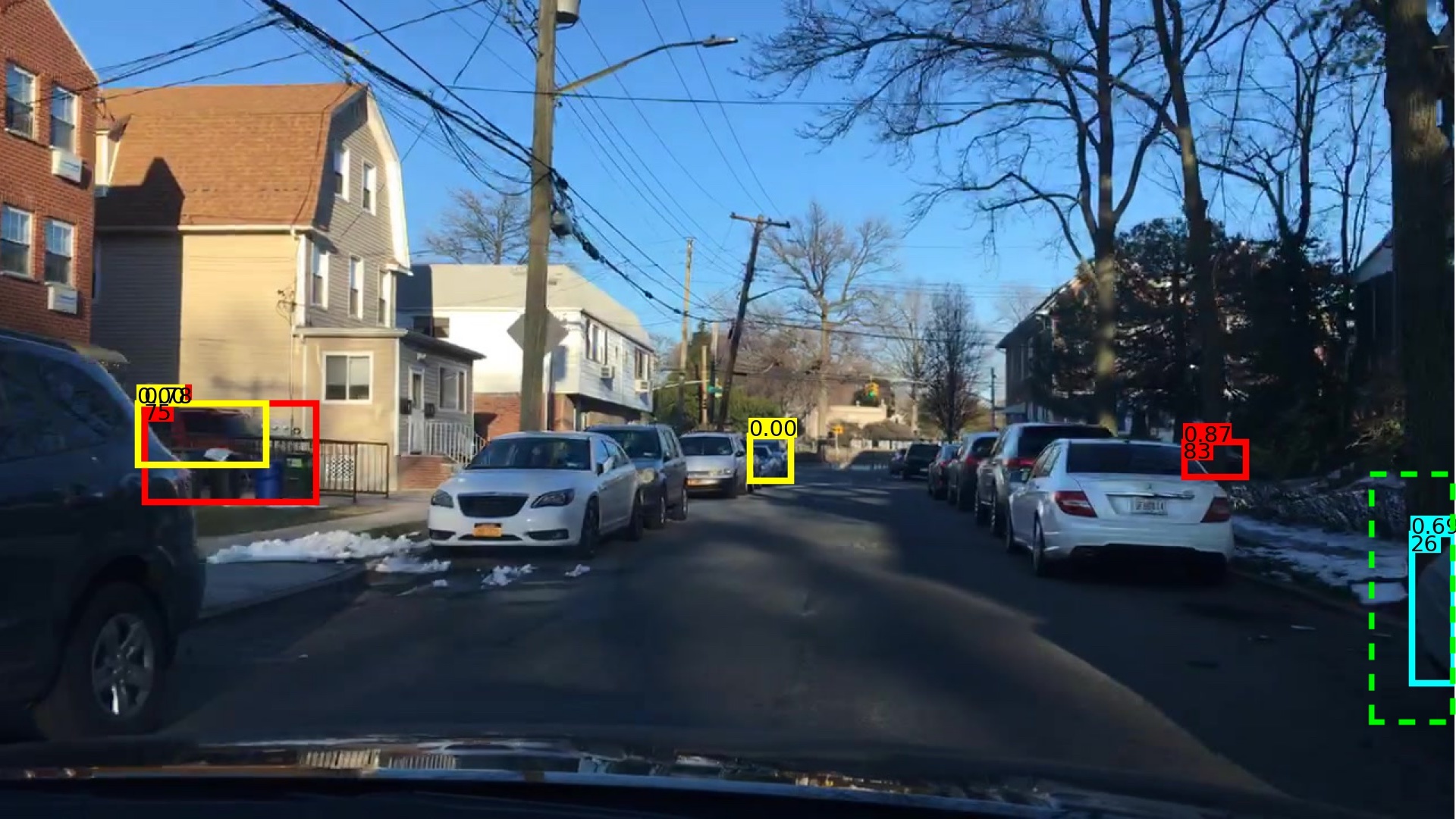}
    \end{subfigure}
    \caption{\textbf{Extreme high-level truncation.} Our method cannot distinguish different instances effectively according to the limited appearance information in highly truncated objects.}
    \label{fig:truncate}
\end{figure*}

Another corner case is the extreme high-level truncation.
As shown in Figure~\ref{fig:truncate}, the highly truncated objects only appear a little when they just enter or leave the camera view.
We cannot distinguish different instances effectively according to the limited appearance information.

\begin{figure*}
    \centering
    \includegraphics[width=\linewidth]{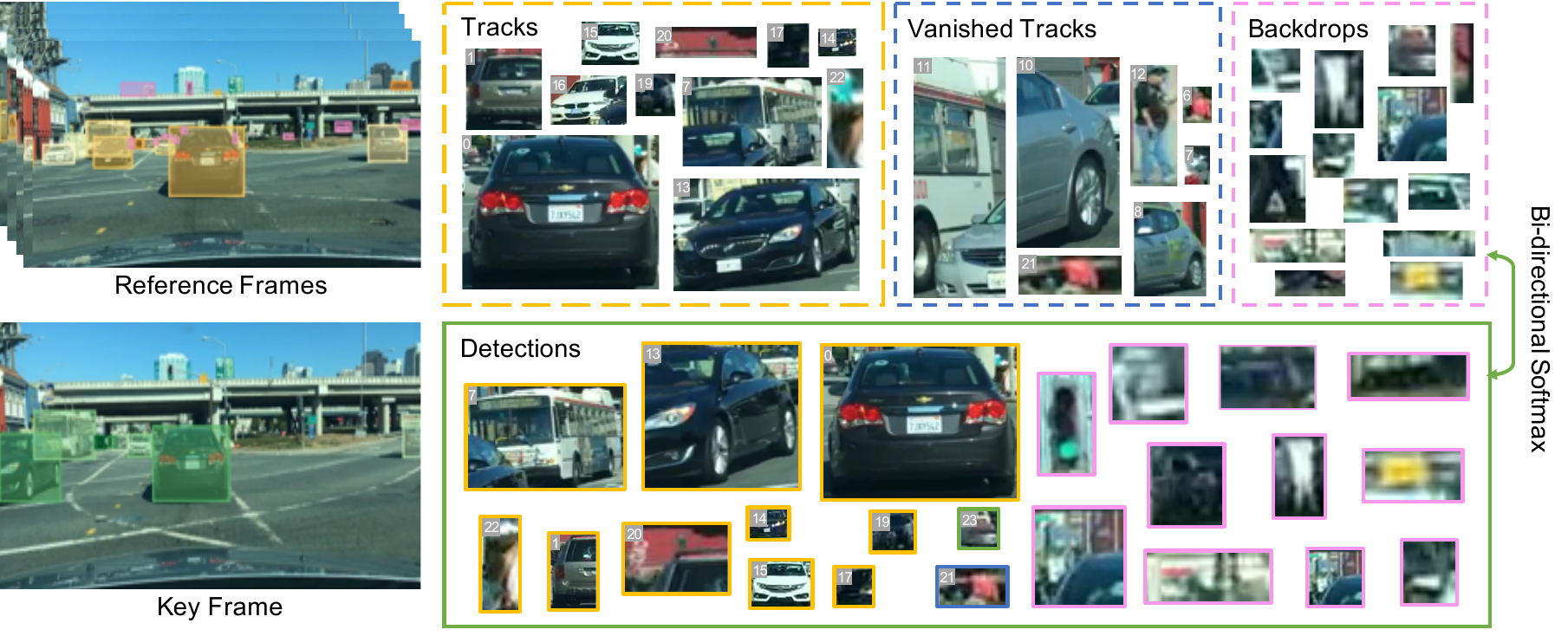}
    \caption{\textbf{Visualizations of different instance patches during inference.} The detected objects in the current frame are matched to tracklets in the consecutive frame, vanished tracklets, and backdrops via bi-directional softmax.}
    \label{fig:test_vis}
\end{figure*}

\section{Visualizations}
We show the visualizations of different instance patches during the testing procedure in Figure~\ref{fig:test_vis}.
The detected objects in each frame are matched to prior objects via bi-directional softmax.
The prior objects include tracks in the consecutive frame, vanished tracks, and backdrops.
We annotate them with different colors.
Each detected object is enclosed by the same color of its matched object.
We can observe that most false positives in the current frame are matched to backdrops, which demonstrates keeping backdrops during the matching procedure helps reduce the number of false positives.

\section{Qualitative results}

\begin{figure*}[h!]
    \centering
    \includegraphics[width=\linewidth]{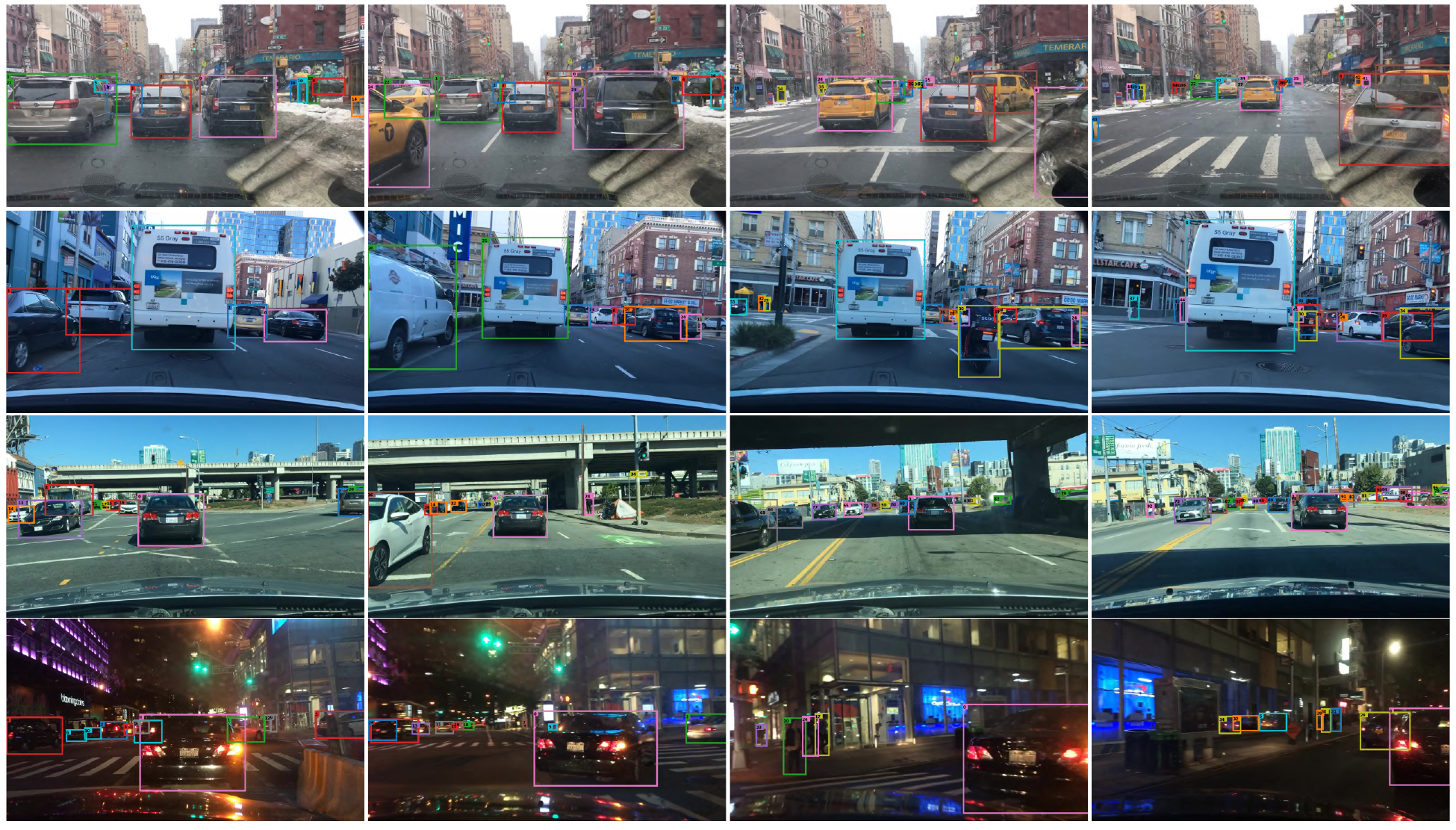}
    \caption{Qualitative results of our method on BDD100K.}
    \label{fig:vis_bdd}
\end{figure*}

\begin{figure*}[h!]
    \centering
    \includegraphics[width=\linewidth]{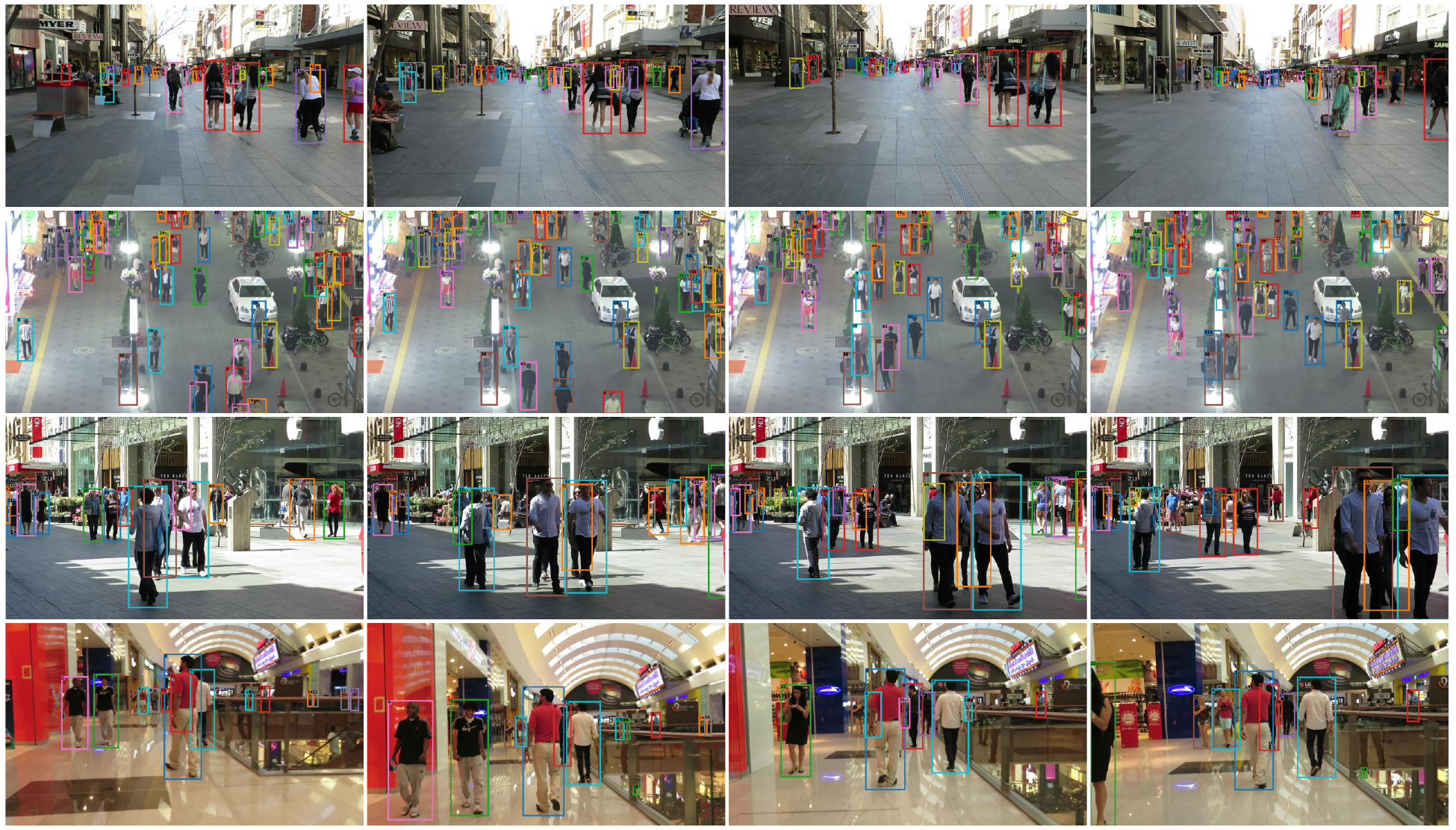}
    \caption{Qualitative results of our method on MOT17.}
    \label{fig:vis_mot}
\end{figure*}

We show some qualitative results of our method on BDD100K dataset and MOT17 dataset in Figure \ref{fig:vis_bdd} and Figure \ref{fig:vis_mot}, respectively. The results are sampled from a certain interval for illustrative purposes.

